\documentclass[sigconf]{acmart}

\AtBeginDocument{%
  }

\copyrightyear{2026}
\acmYear{2026}
\setcopyright{cc}
\setcctype{by-nc-nd}
\acmConference[ICMI '26]{INTERNATIONAL CONFERENCE ON MULTIMODAL INTERACTION}{October 05--09, 2026}{Napoli, Italy}
\acmBooktitle{INTERNATIONAL CONFERENCE ON MULTIMODAL INTERACTION (ICMI '26), October 05--09, 2026, Napoli, Italy}
\acmDOI{10.1145/3776574.3831151}
\acmISBN{979-8-4007-2318-6/2026/10}

\begin{document}

\title{Speech Signals Complement LLMs for Predicting Interpersonal Attraction in Speed Dating}

\author{Yuriko Kikuchi}
\orcid{0009-0009-0097-2508}
  \affiliation{%
    \institution{Japan Advanced Institute of Science and Technology}
    \city{Nomi}
    \state{Ishikawa}
    \country{Japan}
}
\email{yuriko\_kikuchi@jaist.ac.jp}

\author{Takato Hayashi}
\orcid{0009-0000-9757-9431}
\affiliation{%
    \institution{Japan Advanced Institute of Science and Technology}
    \city{Nomi}
    \state{Ishikawa}
    \country{Japan}}
\email{hayashi0884@jaist.ac.jp}

\author{Ryusei Kimura}
\orcid{0009-0006-2254-7228}
\affiliation{%
    \institution{Japan Advanced Institute of Science and Technology}
    \city{Nomi}
    \state{Ishikawa}
    \country{Japan}}
\email{ryusei\_kimura@jaist.ac.jp}

\author{Naoya Inoue}
\orcid{0000-0002-2961-6939}
\affiliation{%
    \institution{Japan Advanced Institute of Science and Technology}
    \city{Nomi}
    \state{Ishikawa}
    \country{Japan}}
\email{naoya-i@jaist.ac.jp}

\author{Ryo Ishii}
\orcid{0009-0001-3849-1656}
\affiliation{%
  \institution{NTT, Inc.}
  \city{Yokosuka}
  \state{Kanagawa}
  \country{Japan}}
\email{ryoct.ishii@ntt.com}

\author{Shogo Okada}
\orcid{0000-0002-9260-0403}
\affiliation{%
    \institution{Japan Advanced Institute of Science and Technology}
    \city{Nomi}
    \state{Ishikawa}
    \country{Japan}}
\email{okada-s@jaist.ac.jp}

\renewcommand{\shortauthors}{Kikuchi et al.}

\begin{abstract}
Large language models (LLMs) can predict interpersonal attraction from conversation transcripts, but it remains unclear what a speech predictor can add beyond transcript-only LLM prediction. Using Japanese speed-dating conversations, we combine predictions from a transcript-only LLM and a supervised speech predictor to estimate participants' reported liking of their partners. We show that speech can complement transcript-only LLM prediction, but that this complementarity is conditional rather than universal. Combining the two predictions significantly improves pairwise ranking accuracy over the transcript-only LLM alone in all evaluated conditions. By contrast, gains in per-participant Pearson $r$ vary across conversation rounds and rating directions, with none significant after correction. Retrospectively, these $r$ gains are concentrated among participants for whom the speech predictor is more accurate. Speech can therefore retain predictive value even when an LLM predicts attraction from transcripts. The relevant question is not simply whether speech helps, but where its complementarity emerges.\footnote{Code is available at \href{https://github.com/yurikomium/speech-llm-complementarity}{github.com/yurikomium/speech-llm-complementarity}.}
\end{abstract}




\maketitle
\section{Introduction}
\label{sec:introduction}

Interpersonal attraction formed during an initial encounter can shape whether a relationship continues to develop~\cite{Berscheid1998Attraction}. Speed dating provides a naturalistic setting for studying how initial attraction forms: participants meet multiple potential partners and report how much they like each one. Conversational behavior can be informative in this setting. Pitch convergence, for example, varies with how speakers perceive their partners' visual attractiveness and overall likability~\cite{MichalskySchoormann2017}. These findings motivate computational models of attraction from interaction data.

Automatic prediction of interpersonal outcomes from conversations has developed along two largely separate lines. One uses large language models (LLMs) to infer outcomes from transcripts. In a direct study of speed dates, LLMs predicted objective and subjective indicators of interaction success from conversation transcripts~\cite{Matz2025Attraction}. The other uses supervised models with lexical, speech, dialogue, and visual signals to predict social judgments and outcomes~\cite{Ranganath2009EMNLP,Veenstra2011ICCVW}. Together, these lines raise a broad question about which conversational signals remain useful beyond transcript-only LLM prediction. We address a narrower question for speed-dating attraction: can a supervised speech predictor add predictive value beyond a transcript-only LLM, and where does that value emerge?

We study this question using the Multi-Modal Speed Dating corpus~\cite{Ishii2023INTERACT}, which contains Japanese speed-dating conversations with transcripts, speaker-specific audio, and post-conversation liking scores. Our task is to predict each participant's reported liking of their partners, with an emphasis on differentiation within the participant's partner set. We define \emph{speech-derived complementarity} as the additional predictive value contributed by a supervised speech predictor beyond the evaluated transcript-only LLM's predictions, without claiming that the speech predictor captures information fundamentally absent from transcripts.

\begin{figure*}[t]
  \includegraphics[width=\textwidth,trim=320 286 24 326,clip]{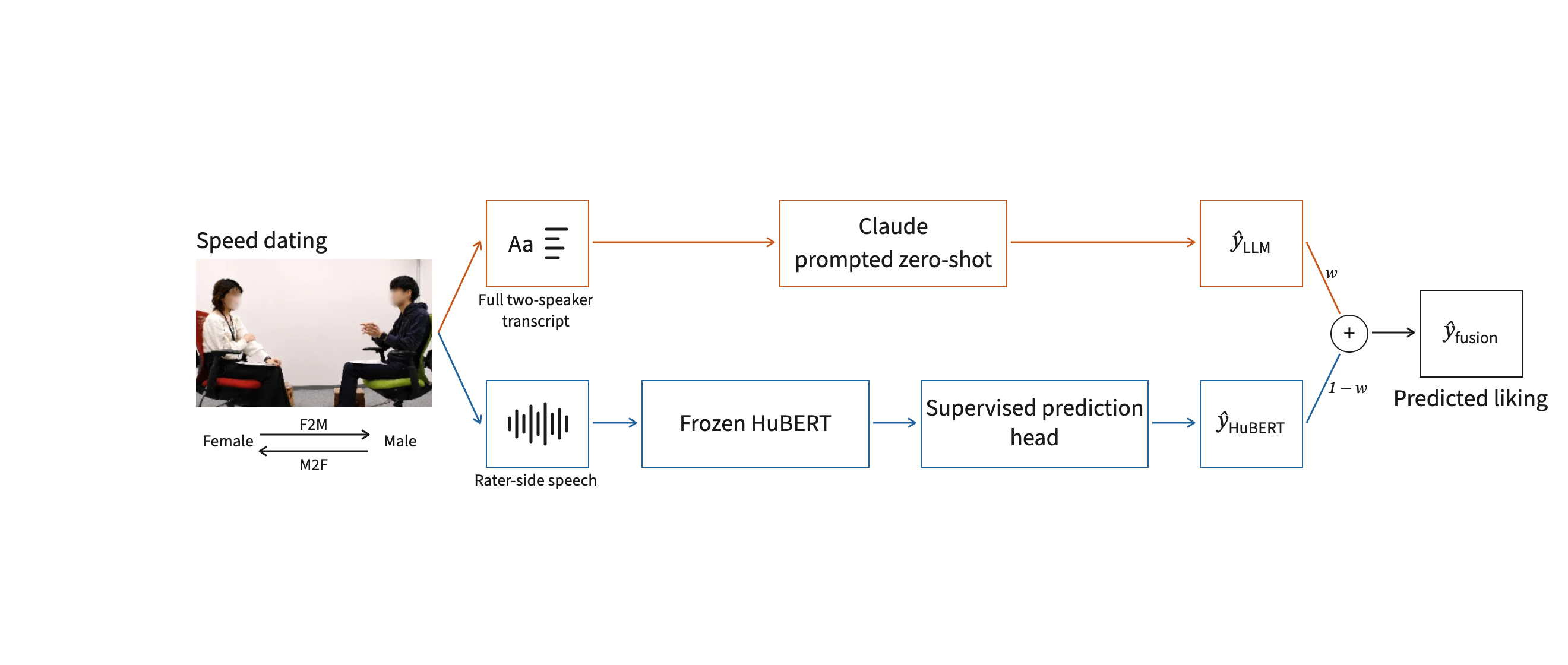}
  \caption{Prediction pipeline for F2M (female rating male) and M2F (male rating female). Claude Sonnet~4.6, used with extended thinking disabled and zero-shot prompting, predicts liking from the full two-speaker transcript, while frozen HuBERT and a supervised head predict from rater-side speech. Weighted score-level fusion combines the predictions using a fold-selected weight $w$.}
  \Description{A prediction pipeline for liking after a speed-dating conversation. F2M denotes a female participant rating a male partner, and M2F denotes a male participant rating a female partner. The conversation feeds two branches. The upper branch sends the full two-speaker transcript to Claude Sonnet 4.6 with zero-shot prompting and extended thinking disabled, yielding the transcript-only prediction. The lower branch sends rater-side speech through frozen HuBERT and a supervised prediction head, yielding the HuBERT prediction. The two predictions are combined using weights $w$ and $1-w$ to produce the fused predicted liking score.}
  \label{fig:pipeline}
\end{figure*}

For the primary analysis, the transcript-only predictor is Claude Sonnet~4.6 with extended thinking disabled (hereafter Claude), evaluated using zero-shot prompting. The supervised speech predictor is built on frozen HuBERT-Large representations (hereafter HuBERT). We combine their scalar predictions through weighted score-level late fusion, as shown in Figure~\ref{fig:pipeline}. This deliberately simple combination serves as a diagnostic for additional predictive value rather than as a new fusion architecture. We evaluate complementarity across two conversation rounds, two rating directions, and participants. Two multimodal LLMs that receive transcripts and audio provide auxiliary direct-input comparisons.

In summary, we clarify where a supervised speech predictor adds predictive value beyond the evaluated transcript-only LLM in speed-dating attraction prediction. We do so by examining this added value across evaluation metrics, conversation rounds and rating directions, and participants, using score-level late fusion only as a diagnostic. Supporting analyses assess whether the transcript-only LLM and speech predictor make redundant predictions by quantifying their within-rater prediction overlap and mutual incremental associations. Together, these analyses shift the question from whether speech helps on average to where its complementarity emerges.

Relative to Claude, fusion significantly improves pairwise ranking accuracy in all four round-by-rating-direction settings. Gains in per-participant Pearson $r$, however, vary across these settings, and no paired gain is statistically significant after correction. Retrospectively, participant-level $r$ gains are larger where HuBERT is more accurate, although this association is partly expected because the HuBERT prediction enters the fused score. The Claude and HuBERT predictions share little within-rater variance, with each explaining comparable incremental in-sample variance beyond the other. As supporting context, Claude has higher per-participant $r$ than every tested supervised text model in all four settings.

Taken together, these findings do not support a blanket claim that adding speech uniformly improves attraction prediction. Instead, they show that speech can retain predictive value even when an LLM predicts attraction from transcripts. For multimodal social prediction, the relevant question is therefore not simply whether speech helps, but where its complementarity emerges.

\section{Related Work}
\label{sec:related}

\subsection{LLMs for Interpersonal Inference}

Beyond attraction, LLMs have been evaluated as judges of empathic communication in text-based conversations. When given criteria for each framework and expert-annotated examples, LLM--expert agreement approached expert--expert agreement across four evaluative frameworks~\cite{Kumar2026Empathy}.

Most directly related to the present paper, Matz et al.~\cite{Matz2025Attraction} applied ChatGPT to speed-dating transcripts to predict mutual contact exchange as well as subjective outcomes and experiences. ChatGPT's correlations with these criteria were modest ($r = .12$--$.23$). For mutual contact exchange specifically, its performance was comparable to that of transcript-only human judges ($r = .12$ vs.\ $.13$). Building on this work, we test whether a supervised speech predictor adds predictive value beyond transcript-only LLM predictions for continuous liking and within-participant partner ranking.

\subsection{Multimodal Prediction of Interpersonal Outcomes}

Interpersonal outcomes have long been studied computationally using verbal and nonverbal features from conversation. Brief observations of expressive behavior can support accurate interpersonal judgments, and nonverbal communication encompasses cues in the face, voice, body, touch, and interpersonal space~\cite{AmbadyRosenthal1992, Hall2019NVC}. In speed dating, supervised models have used prosodic, dialogue, and lexical features to detect flirtation~\cite{Ranganath2009EMNLP} and position, proximity, and motion features from video to predict contact exchange and physical attraction~\cite{Veenstra2011ICCVW}. Complementary statistical analyses have linked conversational behavior to perceived connection in speed-dating encounters~\cite{McFarland2013AJS} and function-word language style matching to mutual romantic interest~\cite{Ireland2011PsychSci}. Speakers' pitch convergence also varies with their perception of the partner's visual attractiveness and overall likability~\cite{MichalskySchoormann2017}. On Japanese speed-dating data, Ishii et al.~\cite{Ishii2023INTERACT} predicted post-interaction liking scores from information available before the interaction, including profiles, facial features, and psychometric measures. More recently, frozen self-supervised speech representations, including HuBERT, have transferred effectively to paralinguistic tasks such as emotion recognition~\cite{Hsu2021HuBERT, yang21c_interspeech}.

Santana et al.~\cite{Santana2025SpeechToJoy} provide the closest architectural precedent for our supervised branches. Their Speech-to-Joy framework trains separate predictors over frozen pretrained text and audio embeddings, using additive attention pooling and modality-specific regression heads. The framework then averages the two predictors' output scores for enjoyment prediction in human--robot dialogue. We adapt this supervised-branch architecture, but our main comparison pairs the speech predictor with a transcript-only LLM prediction rather than a supervised text prediction.

\subsection{LLM Integration in Multimodal Prediction}

Multimodal fusion methods are conventionally partitioned into feature-level early fusion and score-level late fusion. Late fusion is particularly well suited when the predictors are trained under different regimes, since each can be optimized independently. Recent work has integrated LLMs into multimodal prediction pipelines in several ways. Some approaches textualize nonverbal modalities in the prompt~\cite{Ma2025ICMI, hasan2023textmitextualizemultimodalinformation} or use an LLM for post hoc refinement of conventional model outputs~\cite{Singh2025Interspeech}. Others use an LLM to integrate modality-specific predictions in natural language rather than through a weighted score-level combination~\cite{Demirel2025AppleActivity}.

At the task level, Pereira et al.~\cite{Pereira2024ICMI} combined an LLM-derived enjoyment score with temporal, visual, and acoustic features in supervised models for human--robot conversations, and separately compared text-only and video-plus-audio LLM inputs. Their supervised target was third-party enjoyment annotations. The LLM score entered the supervised model alongside modality descriptors rather than being fused at score level with the output of a separately trained modality-specific predictor.

Two prior works more specifically combine score-level late fusion with LLMs, but neither treats a transcript-only LLM prediction as an independent branch. In Amin et al.~\cite{Amin2023}, ChatGPT serves as an auxiliary text generator whose output is featurized alongside the raw text, so all fused branches take text input. In ZS-Fuse~\cite{Kataria2026ZSFuse}, whose ``zero-shot large-scale model plus supervised specialist model'' structure parallels ours, the zero-shot branch is itself audio-driven rather than transcript-driven. A separate direction integrates audio directly into LLMs: audio-native models such as SALMONN~\cite{Tang2024SALMONN} integrate pre-trained speech and audio encoders with an LLM and let it reason over audio inputs directly. We include this direction as an auxiliary comparison using the MLLM baselines Gemini 2.5 Flash and GPT-audio-mini with text-plus-audio input.

Together, these studies leave unresolved whether a separately trained speech predictor improves interpersonal-attraction prediction beyond a transcript-only LLM. We treat a transcript-only LLM prediction obtained through zero-shot prompting and a separately trained supervised speech predictor as independent branches, combining their outputs through weighted score-level late fusion. We use this conventional fusion method to diagnose the speech predictor's additional predictive value beyond the transcript-only prediction. Audio-native MLLMs form a separate direct-input route, which we include only as an auxiliary comparison in the present study.

\section{Corpus and Task Definition}
\label{sec:dataset}
\subsection{Speed-Dating Corpus}
\label{sec:corpus}
We use the Multi-Modal Speed Dating corpus~\cite{Ishii2023INTERACT}. It was collected in the speed-dating paradigm, a methodologically validated setting for observing initial romantic attraction under controlled yet ecologically valid conditions~\cite{Finkel2007SpeedDating}. The corpus records face-to-face conversations between male-female pairs involving 147 native Japanese speakers: 75 female and 72 male participants. Their ages ranged from 19 to 60 years, with a mean of 31.9 years and a standard deviation of 8.6 years. All participants had indicated a strong interest in meeting future partners.

Participants were assigned to age-balanced groups of 5 female and 5 male members. Each group held all 25 possible male-female pairwise conversations; some participants joined multiple groups. The corpus is organized into two conversation rounds with nominal durations of 5 minutes (Session~1) and 10 minutes (Session~2). Each session contains 624 pairwise conversations for a total of 1{,}248. The same pairs conversed in both rounds, and Session~2 was held after Session~1. Thus, each Session~2 conversation followed a prior encounter with the same partner. Each conversation is released with a human-made transcript and per-speaker audio recorded via headset microphones.

\subsection{Liking Scores and Evaluation Conditions}
\label{sec:liking_score}
After each conversation, both participants rated their partner using a Japanese translation of the 13-item Rubin's Liking Scale~\cite{Rubin1970Liking,Ishii2023INTERACT}. Each item used a 1--9 Likert scale, and the scale had a Cronbach's $\alpha$ of $.94$. The prediction target is the \emph{liking score}, defined as the mean of the 13 items. The items span multiple facets of liking, including empathy, adaptability, trust in judgment, likability, perceived similarity, intelligence, and admiration.

We treat the two rating directions independently. F2M denotes predictions of liking scores assigned by female participants to their male partners, and M2F denotes the reverse. In both abbreviations, the direction runs from the rater to the rated partner. Within each conversation group, we exclude a rater's conversations if that rater assigned the same liking score to every partner, because correlation is undefined for that subset. After this exclusion, the Session~1 samples contain 604 F2M pairs from 73 raters and 619 M2F pairs from 72 raters. The Session~2 samples contain 619 pairs in each direction, from 75 female raters for F2M and 72 male raters for M2F. Crossing the two conversation rounds with the two rating directions yields the four conditions analyzed throughout. Mean liking scores ranged from 4.36 to 4.94 across the four conditions, with standard deviations near 1.5.

\subsection{Task Definition}
\label{sec:task}

Let $\mathcal{D}_i = \{(X_{ij}, y_{ij})\}_{j=1}^{N_i}$ denote rater $i$'s set of conversations across multiple partners, the central unit of analysis. For each partner $j$, $X_{ij}$ denotes the transcript and/or speech information available from the interaction, and $y_{ij} \in [1,9]$ denotes the liking score that rater $i$ assigns to partner $j$ after the conversation. Our task is to develop a predictor $f$ that maps $X_{ij}$ to a scalar score $\hat{y}_{ij} = f(X_{ij})$. The predictor should approximate the absolute liking score ($\hat{y}_{ij} \approx y_{ij}$) while preserving the within-participant ordering of non-tied partner pairs. Preserving this ordering means that, for each non-tied pair within a participant's partner set, the predicted and observed score differences have the same sign.

Recovering within-participant differentiation is our primary inferential target: it captures which partners a participant reported liking more relative to their own baseline, independent of between-participant differences in response-scale use. Absolute score agreement nevertheless remains relevant when predictions are interpreted on the original 1--9 scale, motivating a calibration-sensitive optimization objective. The supervised predictors are therefore trained on the concordance correlation coefficient (CCC)~\cite{Lin1989CCC}. CCC rewards association only when the predicted and observed scores also agree in location and scale.

\section{Models, Fusion, and Evaluation Setup}
\label{sec:modeling}

The primary comparison combines a transcript-only LLM prediction from the full two-speaker transcript with a supervised prediction from rater-side speech. We use conventional weighted score-level late fusion to diagnose the speech predictor's additional value, not as a new architecture. The supervised branches adapt Santana et al.'s architecture~\cite{Santana2025SpeechToJoy}: frozen representations, additive attention pooling, and modality-specific regression heads.

\subsection{Supervised Single-Modality Predictors}
\label{sec:supervised_predictors}

We train separate supervised text and speech predictors rather than a joint encoder. Each places a small learnable module on a frozen pretrained encoder, so only the downstream aggregation and regression are learned from the speed-dating corpus. Both predictors share the same downstream architecture: attention pooling over the rater's own utterances, followed by a scalar regression head. Holding the downstream architecture constant helps isolate differences in what the encoders extract from the conversation. It also makes the two predictors directly compatible inputs to score-level late fusion. In what follows, we refer to each supervised predictor by the name of its underlying encoder (e.g., HuBERT denotes the supervised predictor built on frozen HuBERT features, not the encoder alone).

\paragraph{Text embeddings}
Consecutive same-speaker transcript segments are merged before utterance-level encoding. Each utterance is encoded with Sentence-T5-large (335M parameters)~\cite{ni-etal-2022-sentence}, yielding an L2-normalized 768-dimensional sentence embedding.

\paragraph{Speech embeddings}
We use pretrained HuBERT-Large (317M parameters)~\cite{Hsu2021HuBERT} as the speech encoder, a self-supervised speech model pretrained on Libri-Light~\cite{Kahn2020LibriLight}. All parameters are frozen. For each utterance we feed the raw waveform (16 kHz) through HuBERT and average-pool the final hidden states to obtain a 1{,}024-dimensional embedding (L2-normalized). Utterances are cut by transcript timestamps with silence removed via WebRTC voice activity detection.

\paragraph{Architecture details}
The encoder stays frozen throughout training because each cross-validation training split contains only on the order of $10^2$ training conversations, a regime in which fine-tuning a $10^8$-parameter model is prone to overfit. Utterance-level embeddings are aggregated by additive attention pooling, letting the model emphasize utterances that carry rater-side signal about liking rather than weighting all utterances equally. Pooling is restricted to the rater's own utterances; prior speed-dating work shows that speakers modulate voice pitch with their own mate preferences, providing precedent for rater-side acoustic signal~\cite{Pisanski2018}. The pooled representation passes through a two-layer head. A linear layer projects it to hidden dimension $\max(\lfloor d/5 \rfloor, 16)$ (where $d$ is the encoder output, 768 for text and 1024 for audio), followed by ReLU and dropout 0.2; a final linear layer maps it to a scalar. Additional supervised baselines---BERT~\cite{devlin-etal-2019-bert}, the Japanese Linguistic Inquiry and Word Count (J-LIWC)~\cite{Igarashi2022JLIWC}, and openSMILE~\cite{Eyben2010OpenSMILE}---use the same architecture, with $d$ set to each feature's dimension.

\paragraph{Training settings}
All supervised models use CCC loss ($1 - \overline{\text{CCC}}$, where $\overline{\text{CCC}}$ is the mean per-participant CCC across raters in a batch). We optimize them with Adam (learning rate $10^{-3}$, no weight decay or learning-rate scheduling). Batches include up to six raters and all of their conversations. Training runs for a maximum of 500 epochs with early stopping (patience 20 on validation CCC) and random seed 42. Targets are standardized using the training-fold mean and standard deviation; predictions are inverse-transformed and clipped to the valid $[1,9]$ range.

\subsection{LLM-Based Prediction}
\label{sec:llm_zero_shot}
Transcript-only LLM predictions provide the comparison point for testing the speech predictor's additional value. We obtain them using zero-shot prompting, without task-specific fine-tuning or in-context examples. Each model receives the complete chronological transcript, including utterances from both the rater and the partner, with speaker labels and timestamps; no rater-side filtering is applied. The prompt identifies the rater (female for F2M, male for M2F). Adapting Santana et al.'s item-wise questionnaire prediction strategy~\cite{Santana2025SpeechToJoy} to Rubin's Liking Scale, the model predicts the rater's response to each of the scale's 13 items on a 1--9 integer scale using each API's structured-output mechanism. The mean is the predicted liking score ($\in [1, 9]$). This input scope differs from the supervised text and speech branches, whose pooling is restricted to the rater's own utterances. We use temperature 0 for all configurations reported in the main analysis. The prompt structure, Japanese and English versions, and placeholder substitutions are documented in Section~S2.4\footnote{In section, table, and figure numbers, ``S'' denotes the Supplementary Appendix.}; the released code contains the exact executable templates, including the full item wording.

\paragraph{Transcript-only LLMs}
We evaluate proprietary LLMs (Claude Sonnet 4.6 and GPT-5.4) and an open-weight model (Gemma~3 12B-IT) in a transcript-only (T) input configuration, where each model receives this complete two-speaker transcript without audio. The selected Claude configuration disables extended thinking.

\paragraph{Multimodal LLMs}
We also evaluate an alternative route that feeds audio directly to the language model alongside the transcript. As auxiliary baselines, we use Gemini 2.5 Flash and GPT-audio-mini. Each model receives the full transcript and the full conversation audio in a mixed-mono input configuration created from the two headset recordings. Gemini receives lossless FLAC at 16 kHz, whereas GPT-audio-mini receives 128 kbps MP3 passed as base64. For Gemini, we report matched transcript-only (T) and text-plus-audio (T+A) input configurations. The two prompts are identical in transcript content, system role, and item-wise scoring instruction, so the within-model difference isolates the addition of the audio stream under this input configuration. For GPT-audio-mini, we report the T+A configuration as a direct-input baseline and compare it with the transcript-only LLM baselines, but do not make a within-model audio ablation claim. Here, transcript-only LLM and direct-input MLLM describe the evaluated input configurations, not a model's underlying multimodal capability.

\subsection{Representative Selection and Score-Level Late Fusion}
\label{sec:late_fusion}
We perform score-level fusion of predictions from two unimodal representatives, one per input branch. Score-level fusion is necessary for Claude because the branch exposes only a scalar score, not an intermediate embedding, making feature-level concatenation inapplicable. The supervised unimodal predictors likewise produce per-conversation scalar predictions. For the supervised text+speech pair (Sentence-T5 + HuBERT), where feature-level concatenation is technically feasible, Table~\ref{tab:main_results} reports score-level late and utterance-level early fusion as reference points. For the main fusion, we choose one representative per input branch \emph{a priori}, before inspecting any results: Claude for transcript-only prediction and HuBERT for speech. Claude permits the common temperature-0 setting, unlike extended-thinking mode. The broader comparison pool additionally includes BERT, openSMILE, and J-LIWC as supervised baselines, alongside Sentence-T5 and the evaluated LLM and MLLM variants. The retrospective complementarity landscape of this pool (Figure~S1) is consistent with this choice: across all four conditions, HuBERT occupies a favorable region of relatively high utility and low prediction correlation with Claude.

Given predictions $\hat{y}_1$ and $\hat{y}_2$ from the two branch representatives, the fused prediction is
\begin{equation}
\hat{y}_\text{fusion} = w\, \hat{y}_1 + (1-w)\, \hat{y}_2, \quad 0 \le w \le 1.
\end{equation}
For each test fold, the weight $w$ is selected by a 21-point grid search over $w \in \{0, 0.05, \dots, 1.0\}$ to maximize per-participant CCC over the pooled held-out predictions of the remaining folds. The selected weight is then applied to the held-out test fold without further tuning. Because participants can appear in multiple conversation groups, the primary weight-selection pool is not strictly participant-disjoint; excluding all pairs involving test-fold participants yields the same qualitative and inferential conclusions (Section~S2.3).

\subsection{Evaluation Procedure}
\label{sec:evaluation_metrics}
We evaluate predictions by 25-fold Leave-One-Group-Out cross-validation, holding out each of the 25 conversation groups as the test fold. For the supervised models, all pairs involving participants of the held-out group are removed from training, so the trained predictor never sees a test-fold participant. A separate validation fold, participant-disjoint from the test fold, is used for early stopping. Raters with fewer than three partners or near-constant ratings are additionally excluded from the training set only.

For each participant, we compute Pearson's $r$ between predicted and true liking scores across their partners, then average $r$ across participants. Per-participant $r$ is the primary evaluation metric because our central research question concerns whether a model recovers each participant's relative differentiation among partners. Its invariance to positive affine transformations factors out participant-specific location and scale use~\cite{Cronbach1955Accuracy}. Optimization and primary evaluation use different criteria. CCC is calibration-sensitive: unlike $r$, it penalizes differences in prediction mean and variance. We therefore use CCC for both supervised training and fusion-weight selection, discouraging models or weights that preserve association while producing poorly calibrated scores on the original 1--9 scale. We compute CCC per participant and average it across participants, reporting it as a secondary measure of absolute agreement (Table~S2). We also report per-participant Top-1 accuracy and pairwise accuracy (PW), the fraction of same-participant partner pairs whose predicted order matches ground truth. For Top-1, all partners tied at the ground-truth maximum are valid; predicted ties receive fractional credit equal to the proportion belonging to that set. For PW, pairs with tied ground-truth liking are excluded and tied predictions are counted as incorrect. PW depends only on ordering and is therefore invariant to strictly increasing transformations.

\section{Results}
\label{sec:results}

\subsection{Individual Models and Fusion Performance}
\label{sec:results_i}

Table~\ref{tab:main_results} reports $r$ for four transcript-only LLMs, two text-plus-audio MLLMs, five supervised unimodal models, and three fusion variants: two score-level late-fusion variants and one utterance-level early-fusion variant.

\begin{table}[!t]
\centering
\caption{Per-participant Pearson correlation ($r$) for transcript-only LLMs, direct-input text-plus-audio MLLMs, supervised unimodal predictors, and fusion variants. S1/S2 denote Sessions~1/2; F2M denotes female participants rating male partners, and M2F the reverse.}
\label{tab:main_results}
\Description{Rows compare transcript-only LLMs, direct-input text-plus-audio MLLMs, supervised text and speech predictors, supervised fusion, and the primary LLM-plus-speech fusion; columns give per-participant r for the four session-by-direction conditions.}
\small
\setlength{\tabcolsep}{2.25pt}
\renewcommand{\arraystretch}{1.05}
\begin{tabular}{@{}lcccc@{}}
\toprule
Model / configuration & S1 F2M & S1 M2F & S2 F2M & S2 M2F \\
\midrule
\textit{LLMs (T)} & & & & \\
 Claude              & $.248{\scriptscriptstyle \pm .048}$ & $.163{\scriptscriptstyle \pm .044}$ & $.244{\scriptscriptstyle \pm .050}$ & $.238{\scriptscriptstyle \pm .044}$ \\
 Gemini 2.5 Flash    & $.223{\scriptscriptstyle \pm .055}$ & $.183{\scriptscriptstyle \pm .048}$ & $.154{\scriptscriptstyle \pm .053}$ & $.222{\scriptscriptstyle \pm .044}$ \\
 GPT-5.4             & $.256{\scriptscriptstyle \pm .045}$ & $\mathbf{.203}{\scriptscriptstyle \pm .054}$ & $.196{\scriptscriptstyle \pm .048}$ & $.237{\scriptscriptstyle \pm .043}$ \\
 Gemma 3 12B-IT     & $.060{\scriptscriptstyle \pm .050}$ & $.122{\scriptscriptstyle \pm .047}$ & $.052{\scriptscriptstyle \pm .053}$ & $.027{\scriptscriptstyle \pm .048}$ \\
\addlinespace[2pt]
\textit{MLLMs (T+A)} & & & & \\
 Gemini 2.5 Flash    & $.162{\scriptscriptstyle \pm .051}$ & $.116{\scriptscriptstyle \pm .050}$ & $.168{\scriptscriptstyle \pm .041}$ & $.102{\scriptscriptstyle \pm .044}$ \\
 GPT-audio-mini      & $.044{\scriptscriptstyle \pm .044}$ & $.135{\scriptscriptstyle \pm .048}$ & $.149{\scriptscriptstyle \pm .049}$ & $.193{\scriptscriptstyle \pm .045}$ \\
\midrule
\textit{Supervised text predictors} & & & & \\
 Sentence-T5         & $.057{\scriptscriptstyle \pm .059}$ & $.014{\scriptscriptstyle \pm .053}$ & $.101{\scriptscriptstyle \pm .051}$ & $.018{\scriptscriptstyle \pm .052}$ \\
 BERT                & $.044{\scriptscriptstyle \pm .057}$ & $.076{\scriptscriptstyle \pm .042}$ & $-.001{\scriptscriptstyle \pm .049}$ & $.051{\scriptscriptstyle \pm .052}$ \\
 J-LIWC              & $.121{\scriptscriptstyle \pm .042}$ & $.031{\scriptscriptstyle \pm .050}$ & $.046{\scriptscriptstyle \pm .057}$ & $-.077{\scriptscriptstyle \pm .048}$ \\
\addlinespace[2pt]
\textit{Supervised speech predictors} & & & & \\
 HuBERT     & $.105{\scriptscriptstyle \pm .052}$ & $.092{\scriptscriptstyle \pm .054}$ & $.212{\scriptscriptstyle \pm .048}$ & $.183{\scriptscriptstyle \pm .043}$ \\
 openSMILE  & $\dagger$ & $\dagger$ & $.006{\scriptscriptstyle \pm .061}$ & $.088{\scriptscriptstyle \pm .051}$ \\
\addlinespace[2pt]
\textit{Supervised text + speech fusion} & & & & \\
 Sentence-T5 + HuBERT (L) & $.103{\scriptscriptstyle \pm .052}$ & $.097{\scriptscriptstyle \pm .055}$ & $.201{\scriptscriptstyle \pm .048}$ & $.180{\scriptscriptstyle \pm .043}$ \\
 Sentence-T5 + HuBERT (E) & $.106{\scriptscriptstyle \pm .050}$ & $.029{\scriptscriptstyle \pm .049}$ & $.173{\scriptscriptstyle \pm .049}$ & $.147{\scriptscriptstyle \pm .046}$ \\
\midrule
\textit{LLM + speech fusion} & & & & \\
 Claude + HuBERT (L) & $\mathbf{.281}{\scriptscriptstyle \pm .049}$ & $.164{\scriptscriptstyle \pm .046}$ & $\mathbf{.323}{\scriptscriptstyle \pm .044}$ & $\mathbf{.248}{\scriptscriptstyle \pm .044}$ \\
\bottomrule
\end{tabular}
\par\vspace{5pt}
\begin{minipage}{\linewidth}
\footnotesize
\raggedright
\textit{Note.} Values are mean per-participant $r \pm$ standard error of the mean (SE); undefined $r$ values are excluded. \textbf{Bold} marks the highest point estimate in each column. $^\dagger$ denotes $r$ undefined for openSMILE in Session~1 because of near-constant predictions. L and E denote weighted score-level late fusion and utterance-level early fusion, respectively.

\textit{Models.} Claude: Claude Sonnet 4.6 with extended thinking disabled; Sentence-T5: Sentence-T5-large; HuBERT: HuBERT-Large.
\end{minipage}
\end{table}

\paragraph{LLMs and auxiliary MLLM comparisons.}
The best-performing LLM configuration in each condition is transcript-only: Claude leads in Session~2 and GPT-5.4 leads in Session~1, although the Session~1 differences are smaller than the SE reported for either model. Under our mixed-mono input configuration, Gemini's T+A input does not consistently improve on its matched transcript-only input, and GPT-audio-mini with T+A input does not surpass the strongest transcript-only baselines.

\paragraph{Supervised baselines and fusion variants.}
Claude has higher per-participant $r$ than each of the three supervised text models, Sentence-T5, BERT, and J-LIWC, in all four conditions (Wilcoxon signed-rank tests; rank-biserial $r_\text{rb} = .24$--$.62$ over the twelve Claude vs.\ supervised-text comparisons). HuBERT is condition-dependent: its performance is closer to Claude's in Session~2 than in Session~1, where it is substantially weaker. The openSMILE predictor produces near-constant outputs for a majority of Session~1 raters. Consequently, per-participant $r$ is undefined on the common rater set; Table~\ref{tab:main_results} marks these results as undefined with $\dagger$. The score-level late fusion of Sentence-T5 and HuBERT is below the Claude--HuBERT fusion in all four conditions, so the Claude prediction is not interchangeable with the Sentence-T5 prediction under the shared fusion setup. The utterance-level early-fusion variant is lower than the score-level variant in three conditions; the two are essentially tied in Session~1 F2M. These results establish Claude as the transcript-only comparison point and HuBERT as a speech predictor whose performance varies by condition.

\subsection{Fusion Analysis}
\label{sec:results_ii}

\paragraph{Condition-dependent fusion gain.}

Throughout, we define fusion gain relative to Claude as $\Delta = \text{Fusion} - \text{Claude}$, using $\Delta r$ for per-participant $r$ and $\Delta$~PW for pairwise accuracy. Relative to Claude, the Claude--HuBERT fusion significantly improves pairwise accuracy in all four conditions, with Holm-corrected $p < .05$ in each (Table~\ref{tab:rank_distance}). By contrast, $\Delta r$ is largest in Session~2 F2M, smaller in Session~1 F2M, and essentially zero in both M2F conditions (Table~\ref{tab:main_results}); no paired $\Delta r$ reaches significance after Holm correction. The fused model also exceeds HuBERT alone in pairwise accuracy in three of four conditions, with Holm-corrected $p < .05$ in each. Session~1 M2F is the exception: HuBERT is the stronger single predictor, and fusion does not improve on HuBERT. The selected weights are comparable between Claude and HuBERT in Session~2 but concentrate on Claude in Session~1 (Figure~S2).

Replacing Claude with GPT-5.4 yields a similar four-condition $\Delta r$ pattern (Table~S7). This robustness does not extend to the tested alternative representatives: substituting Sentence-T5 for Claude or openSMILE for HuBERT lowers fusion per-participant $r$ in every condition (Table~S1). Table~\ref{tab:incremental_variance} further shows little shared within-rater explained variance between Claude and HuBERT, with each providing a similar positive increment beyond the other.

\paragraph{Pairwise accuracy by rank distance.}

Aggregate PW shows that fusion improves pairwise ordering over Claude, but not where the gains occur. Table~\ref{tab:rank_distance} therefore breaks down fusion gains by within-participant rank distance using Near ($1$--$2$), Mid ($3$--$5$), and Far ($6+$) bins, and reports Top-1 change separately. In Session~2 F2M, where the $r$ gain is largest, $\Delta$~PW increases with rank distance; this pattern is less clear elsewhere. Tie-aware Top-1 changes are small, decreasing slightly in Session~1 and increasing slightly in Session~2, so aggregate PW gains do not correspond to uniform improvements in identifying the top-ranked partner.

\begin{table*}[t]
\centering
\caption{Pairwise accuracy (PW) and fusion gains over Claude, reported overall and by rank distance, together with changes in Top-1 accuracy. S1/S2 denote Sessions~1/2; F2M denotes female participants rating male partners, and M2F the reverse.}
\label{tab:rank_distance}
\Description{Rows give the four session-by-direction conditions; columns report aggregate pairwise accuracy for Claude, HuBERT, and fusion, the fusion gain and effect size, rank-distance-specific gains, and the change in Top-1 accuracy.}
\small
\setlength{\tabcolsep}{3pt}
\renewcommand{\arraystretch}{1.05}
\begin{tabular}{lccccccccc}
\toprule
 & \multicolumn{3}{c}{\textit{Aggregate PW}} & \multicolumn{6}{c}{\textit{Fusion comparison with Claude}} \\
\cmidrule(lr){2-4}\cmidrule(lr){5-10}
Condition & Claude & HuBERT & Fusion & Overall PW gain & $r_\text{rb}$ & Near PW gain & Mid PW gain & Far PW gain & Top-1 change \\
\midrule
S1 F2M & .549 & .531 & .603 & $+.054$ & $.594^{***}$ & $+.044$ & $\mathbf{+.076}$ & $+.074$ & $-.025$ \\
S1 M2F & .518 & .560 & .543 & $+.025$ & $.623^{**}$  & $+.026$ & $\mathbf{+.039}$ & $+.031$ & $-.007$ \\
S2 F2M & .579 & .597 & .628 & $+.049$ & $.299^{*}$   & $+.031$ & $+.055$ & $\mathbf{+.091}$ & $+.013$ \\
S2 M2F & .545 & .564 & .594 & $+.049$ & $.374^{*}$   & $+.044$ & $+.042$ & $\mathbf{+.050}$ & $+.028$ \\
\bottomrule
\end{tabular}
\par\vspace{5pt}
\begin{minipage}{\textwidth}
\footnotesize
\raggedright
\textit{Note.} Claude denotes Claude Sonnet~4.6 with extended thinking disabled. Aggregate PW (left) is macro-averaged across participants; rank-distance bins (Near $=$ 1--2, Mid $=$ 3--5, Far $=$ 6+) are micro-averaged over all within-participant pairs (rank distance is a pair-level attribute), so bins need not average to the aggregate. PW gain and Top-1 change are computed as Fusion $-$ Claude. $r_\text{rb}$ is the rank-biserial effect size for the Fusion vs.\ Claude Wilcoxon signed-rank test. \textbf{Bold} marks the bin with the largest PW gain per condition. $^{*}p<.05$, $^{**}p<.01$, $^{***}p<.001$ (Holm-corrected, $m=4$).
\end{minipage}
\end{table*}

\begin{table}[t]
\centering
\caption{Mutual incremental within-rater variance between Claude and HuBERT across the four conditions.}
\label{tab:incremental_variance}
\Description{Rows give Session 1 and Session 2 for female-to-male and male-to-female ratings; columns report the incremental within-rater explained variance of HuBERT beyond Claude, of Claude beyond HuBERT, and their shared explained variance.}
\small
\setlength{\tabcolsep}{4pt}
\renewcommand{\arraystretch}{1.05}
\begin{tabular}{lccc}
\toprule
& \multicolumn{2}{c}{Incremental $R^2$} & \\
\cmidrule(lr){2-3}
Condition & \shortstack{HuBERT beyond\\Claude} & \shortstack{Claude beyond\\HuBERT} & Shared $R^2$ \\
\midrule
S1 F2M & $+.17$\,{\scriptsize$[.13,\,.22]$} & $+.19$\,{\scriptsize$[.15,\,.24]$} & $.04$ \\
S1 M2F & $+.20$\,{\scriptsize$[.15,\,.26]$} & $+.15$\,{\scriptsize$[.11,\,.19]$} & $.01$ \\
S2 F2M & $+.21$\,{\scriptsize$[.16,\,.26]$} & $+.24$\,{\scriptsize$[.19,\,.29]$} & $.01$ \\
S2 M2F & $+.14$\,{\scriptsize$[.11,\,.17]$} & $+.17$\,{\scriptsize$[.13,\,.21]$} & $.03$ \\
\bottomrule
\end{tabular}
\par\vspace{5pt}
\begin{minipage}{\linewidth}
\footnotesize
\raggedright
\textit{Note.} Claude denotes Claude Sonnet~4.6 with extended thinking disabled. ``HuBERT beyond Claude'' is the increase in $R^2$ obtained by adding the HuBERT prediction to the Claude prediction; ``Claude beyond HuBERT'' is defined symmetrically. Shared $R^2$ is the overlap in explained variance. Values are within-rater, in-sample averages (hence the increments are $\ge 0$); brackets show 95\% CIs based on 2{,}000 participant bootstraps.
\end{minipage}
\end{table}

\section{Discussion}
\label{sec:discussion}

\subsection{When Does Speech Complement the Transcript-Only LLM?}

Speech-derived complementarity is conditional across evaluation metrics, conditions, and participants. Relative to Claude, fusion significantly improves PW in all four conditions (Table~\ref{tab:rank_distance}), indicating more accurate ordering of partner pairs. By contrast, no paired $\Delta r$ reaches significance after correction, so the results do not show a uniform improvement in participants' overall differentiation among partners. Absolute per-participant $r$ also remains modest for all models; the gains indicate complementarity, not deployment-ready accuracy.

At the condition level, HuBERT is closer to Claude in Session~2 than in Session~1, and the selected fusion weights are correspondingly more balanced in Session~2 (Figure~S2). Stronger HuBERT performance does not, however, map uniformly onto larger $r$ gains: the largest numerical gain occurs in Session~2 F2M, but the two M2F conditions show essentially no change. HuBERT's stronger Session~2 performance may reflect both prior-encounter exposure and longer within-encounter observation, which this corpus cannot disentangle. Pitch convergence in speed dating varies with perceived partner attractiveness~\cite{MichalskySchoormann2017}, and longer interaction may permit richer partner representations~\cite{AltmanTaylor1973SPT}. The observed condition pattern therefore locates, but does not explain, where complementarity emerges.

\begin{figure*}[t]
\centering
\vspace*{4pt}
\includegraphics[width=\textwidth,trim=0 6 0 14,clip]{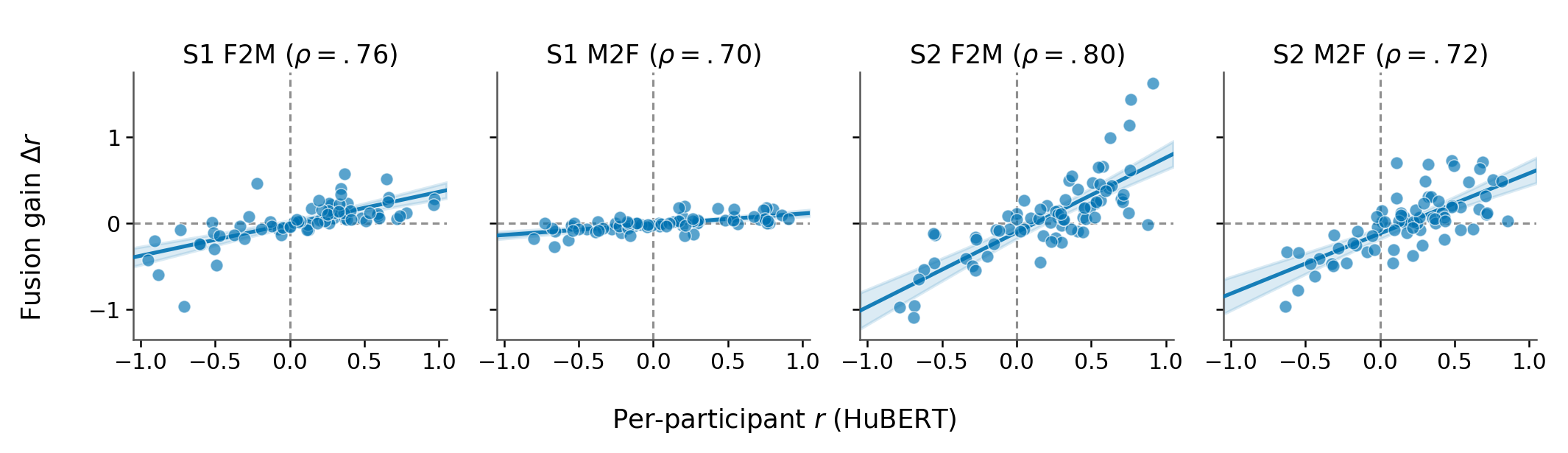}
\caption{Across all four conditions, participants with higher HuBERT per-participant Pearson $r$ show larger Claude--HuBERT fusion gain $\Delta r$ over Claude, although gain magnitude varies by condition. Panels show Sessions~1/2 and female-to-male (F2M)/male-to-female (M2F) ratings. Points are participants; solid lines and bands show OLS fits and 95\% CIs, and dashed lines mark zero. Because HuBERT enters the fused score, this association is descriptive rather than an independent mechanism.}
\Description{Four scatter panels show Session 1 and Session 2 for female-to-male and male-to-female ratings. Each panel plots participants with HuBERT's per-participant Pearson correlation on the horizontal axis and the gain from weighted score-level late fusion of Claude Sonnet 4.6 with extended thinking disabled and HuBERT over Claude alone on the vertical axis. Dashed horizontal and vertical reference lines mark zero. Each panel includes an upward-sloping ordinary least squares line and its shaded 95 percent confidence band. Panel titles report Spearman rank correlations of +0.76 and +0.70 in Session 1 for female-to-male and male-to-female ratings, respectively, and +0.80 and +0.72 in Session 2. All p-values are below .0001, and the panels contain 72 to 75 participants.}
\label{fig:complementarity}
\end{figure*}

\label{sec:complementarity_scatter}
Figure~\ref{fig:complementarity} shows the participant-level relationship between HuBERT's $r$ and fusion gain $\Delta r$. Across participants, Claude's $r$ and HuBERT's $r$ are weakly associated ($|\rho| < .30$ in all conditions), whereas HuBERT's $r$ is strongly associated with $\Delta r$ in every condition ($\rho \geq .70$, all $p < .0001$). Because $\Delta r$ is computed from a fused prediction that already contains the HuBERT score, the correlation between HuBERT's $r$ and $\Delta r$ is partly structurally expected and is not an independent test of a participant-level mechanism. The correlation instead provides a descriptive localization of the observed gains: participants for whom HuBERT is weak also show little fusion benefit. Prior multimodal affect research has emphasized aggregate fusion gains over unimodal baselines~\cite{DMelloKory2015, Santana2025SpeechToJoy}, leaving such participant-level variation largely unexplored.

A within-rater variance decomposition further characterizes the overlap between the Claude and HuBERT predictions (Table~\ref{tab:incremental_variance}). The two predictions are weakly correlated within raters (mean $r$ between $-.08$ and $+.13$), and their shared $R^2$ is at most $.04$ in every condition. The incremental associations are comparable in both directions, indicating mutual rather than one-sided association. The low shared $R^2$ and comparable bidirectional increments are consistent with limited overlap between the two predictions, but the decomposition does not by itself distinguish distinct information from variance reduction under noisy prediction.

\subsection{Late Fusion vs. Direct Multimodal Input}
\label{sec:late_vs_direct}

The auxiliary MLLM comparison examines a different route for incorporating speech. Under the tested mixed-mono configuration, Gemini's text-plus-audio input does not consistently improve on its matched transcript-only input, and GPT-audio-mini does not surpass the strongest transcript-only baselines. By contrast, combining independently obtained Claude and HuBERT predictions yields a pairwise-ordering benefit in the present configuration. The observed contrast between direct multimodal input and score-level fusion does not establish a general advantage of score-level late fusion over direct multimodal inference; the comparison is specific to the evaluated models, prompts, and audio processing.

Score-level fusion instead provides a modular way to evaluate the speech predictor's additional predictive value relative to a transcript-only LLM. The selected weights place less weight on HuBERT in Session~1, where it is less accurate, and relatively more weight on it in Session~2. As transcript-only LLMs evolve, the magnitude of speech-derived complementarity remains an empirical question across models and conditions.

\subsection{Limitations and Future Work}
\label{sec:limitations}

\paragraph{Evaluation target and partner-level consensus}
Per-participant $r$ evaluates within-participant differentiation among partners but does not distinguish rater-specific partner preference from partner-level consensus shared across raters. To contextualize this limitation, we construct a descriptive partner-level consensus reference using the mean score each partner received from other raters. The reference reaches per-participant $r$ of $.36$--$.48$, above the Claude--HuBERT fusion in all four conditions (Section~S1.6). These ground-truth ratings are unavailable at inference, so the reference provides context rather than a deployable baseline.

\paragraph{Mechanism of speech-derived complementarity}
Figure~\ref{fig:complementarity} shows that fusion gains are concentrated among participants for whom HuBERT has higher $r$, but this association is retrospective: it requires all conversations from a participant and cannot predict fusion benefit in advance. Table~\ref{tab:incremental_variance} characterizes the mutual incremental associations between Claude and HuBERT. Replacing HuBERT with the hand-crafted openSMILE representation does not improve on Claude in any condition (Table~S1). These analyses do not identify which acoustic information HuBERT captures. HuBERT represents sequential speech structure~\cite{Hsu2021HuBERT} and transfers well to paralinguistic tasks~\cite{yang21c_interspeech}. Which cues contribute---laughter, pauses, backchannels, or turn-taking---remains open. Future work should test whether participant-adaptive or hierarchical models can prospectively estimate speech-branch reliability.

\paragraph{Direction asymmetry and conversation round}
Numerical $\Delta r$ differs across rating directions and is largest in Session~2 F2M. Because Session~2 combines prior-encounter exposure with a longer nominal duration, this corpus cannot separate their contributions. Future studies should vary these factors independently. Modeling actor and partner effects, interpersonal synchrony, and turn-level dynamics such as prosodic adaptation may also help explain the observed condition dependence.

\paragraph{Corpus, language, and inclusivity}
The evidence comes from a single Japanese speed-dating corpus with 147 participants, restricted to opposite-sex pairs. Whether the condition-dependent complementarity we observe generalizes across languages and pair compositions or to other short, dyadic, first-encounter interactions with an explicit evaluative goal remains an open empirical question.

\paragraph{Model and modality coverage}
The main analysis focuses on text and speech, although video cues can also predict interpersonal outcomes in speed dating~\cite{Veenstra2011ICCVW}. The direct-input MLLM comparison covers two models under one prompting and audio-processing pipeline; audio-LLM capabilities are evolving rapidly~\cite{Sakshi2025MMAU}. We reuse the transcript-only prompt, mix the headset recordings to mono, and supply GPT-audio-mini with compressed MP3 audio. The supervised branches likewise use HuBERT and Sentence-T5 rather than exhaustively comparing alternatives such as wav2vec 2.0~\cite{Baevski2020Wav2Vec2}, WavLM~\cite{Chen2022WavLM}, or Japanese and multilingual encoders. These choices bound the comparison: the experiments neither establish the general performance of direct-input MLLMs nor determine which speech encoder best captures Japanese paralinguistic cues. Future work should evaluate video-aware extensions, audio-specific prompts, speaker-separated audio, and alternative encoders.

\section{Conclusion}
\label{sec:conclusion}

In Japanese speed dating, we examined whether a supervised speech predictor complements a transcript-only LLM in predicting how much participants reported liking each partner. The results show that speech can complement transcript-only LLM prediction, but not as a uniform multimodal gain. Across both conversation rounds and rating directions, fusing speech and transcript-only LLM predictions significantly improves pairwise ranking accuracy over the transcript-only LLM alone. By contrast, gains in per-participant Pearson $r$ vary across these settings, and none is significant after correction. Retrospectively, these $r$ gains are larger for participants whose liking is more accurately predicted by the speech predictor.

Speech can therefore retain predictive value even when an LLM predicts interpersonal attraction from conversation transcripts. For multimodal social prediction, the relevant question is not simply whether speech helps, but how its complementarity varies across evaluation metrics, conversation settings, and participants.

\section*{Safe and Responsible Innovation Statement}
Inferring interpersonal attraction from conversational data risks nonconsensual profiling (e.g., dating-app scoring or workplace surveillance) and overreliance in high-stakes matchmaking. The speed-dating corpus was collected under institutional ethics approval and written informed consent, with access restricted to contract-based research. The main analysis uses transcripts and audio; supplementary checks also use overhead video. These data remain identifying and sensitive. Predictors may reflect pretraining biases, and findings apply only to Japanese opposite-sex speed dating. Responsible deployment requires explicit consent, bias auditing, human oversight, and validation across broader gender identities and cultures.\footnote{The authors used Claude (Anthropic) for drafting text, verifying numerical results against raw outputs, reviewing code, and suggesting the analysis plan. All study decisions and manuscript content remain the authors' responsibility.}

\begin{acks}
This work was partially supported by JSPS KAKENHI Grant Number 26K03016, JST CREST Grant Number JPMJCR2563, JST CRONOS Grant Number JPMJCS24K7, and AMED under Grant Number JP256f0137001.
\end{acks}

\bibliographystyle{ACM-Reference-Format}
\bibliography{references}

\end{document}


\title{Supplementary Appendix to: Speech Signals Complement LLMs for Predicting Interpersonal Attraction in Speed Dating}

\author{Yuriko Kikuchi}
\orcid{0009-0009-0097-2508}
  \affiliation{%
    \institution{Japan Advanced Institute of Science and Technology}
    \city{Nomi}
    \state{Ishikawa}
    \country{Japan}
}
\email{yuriko\_kikuchi@jaist.ac.jp}

\author{Takato Hayashi}
\orcid{0009-0000-9757-9431}
\affiliation{%
    \institution{Japan Advanced Institute of Science and Technology}
    \city{Nomi}
    \state{Ishikawa}
    \country{Japan}}
\email{hayashi0884@jaist.ac.jp}

\author{Ryusei Kimura}
\orcid{0009-0006-2254-7228}
\affiliation{%
    \institution{Japan Advanced Institute of Science and Technology}
    \city{Nomi}
    \state{Ishikawa}
    \country{Japan}}
\email{ryusei\_kimura@jaist.ac.jp}

\author{Naoya Inoue}
\orcid{0000-0002-2961-6939}
\affiliation{%
    \institution{Japan Advanced Institute of Science and Technology}
    \city{Nomi}
    \state{Ishikawa}
    \country{Japan}}
\email{naoya-i@jaist.ac.jp}

\author{Ryo Ishii}
\orcid{0009-0001-3849-1656}
\affiliation{%
  \institution{NTT, Inc.}
  \city{Yokosuka}
  \state{Kanagawa}
  \country{Japan}}
\email{ryoct.ishii@ntt.com}

\author{Shogo Okada}
\orcid{0000-0002-9260-0403}
\affiliation{%
    \institution{Japan Advanced Institute of Science and Technology}
    \city{Nomi}
    \state{Ishikawa}
    \country{Japan}}
\email{okada-s@jaist.ac.jp}

\renewcommand{\shortauthors}{Kikuchi et al.}
\renewcommand{\thesection}{S\arabic{section}}
\renewcommand{\thesubsection}{\thesection.\arabic{subsection}}
\renewcommand{\thesubsubsection}{\thesubsection.\arabic{subsubsection}}
\renewcommand{\thetable}{S\arabic{table}}
\renewcommand{\thefigure}{S\arabic{figure}}

\ccsdesc[500]{Human-centered computing~Empirical studies in collaborative and social computing}
\ccsdesc[300]{Applied computing~Psychology}
\ccsdesc[300]{Computing methodologies~Natural language processing}
\keywords{interpersonal attraction; speed dating; large language models;\texorpdfstring{\linebreak}{ }
speech processing; multimodal fusion}


\maketitle

Throughout this appendix, F2M denotes female participants rating male
partners, and M2F denotes male participants rating female partners; the
direction runs from the rater to the rated partner. Section, table, and figure
numbers use S as the supplementary prefix. Within table and figure condition
labels, S1 and S2 denote Session~1 (nominally 5 minutes) and Session~2
(nominally 10 minutes), respectively.
Unless otherwise stated, Claude denotes the prespecified transcript-only LLM
used in the primary analysis, Claude Sonnet~4.6 with extended thinking
disabled; HuBERT denotes the supervised speech predictor built on frozen
HuBERT-Large representations. The main fusion is their weighted score-level
late fusion. Claude (thinking) denotes the auxiliary Claude configuration with
extended thinking enabled.
Here, LLM denotes a large language model and MLLM denotes a multimodal LLM.
Bare LLM/LLMs refer to the model class or a set of candidate models, not to
Claude as a fixed shorthand.

\section{Additional Evaluation and Robustness}
\label{sec:app_evaluation_robustness}

\subsection{Representative Selection and Substitution Analyses}
\label{sec:app_substitution_ablation}

\begin{figure*}[t]
\centering
\includegraphics[width=\textwidth]{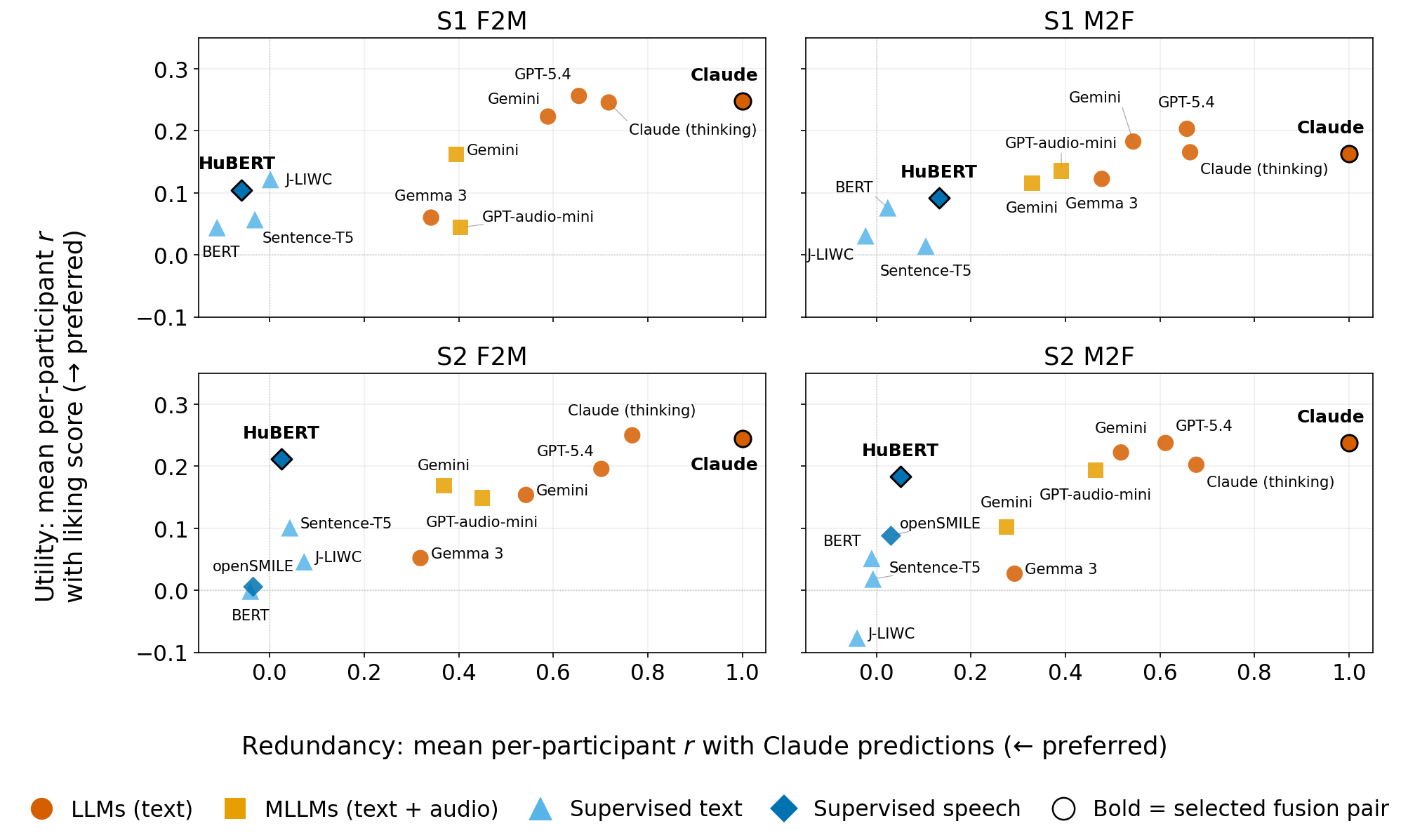}
\caption{Complementarity landscape of candidate models across the four conditions. Each point gives the mean across participants of Pearson's $r$ between a candidate model's predictions and ground-truth liking (utility, y-axis) or the prespecified transcript-only LLM's predictions (Claude; redundancy proxy, x-axis). A favorable speech complement lies upper-left (high utility, low prediction correlation). The selected pair---Claude for transcript-only prediction and HuBERT for supervised speech prediction---is highlighted. openSMILE appears only in Session~2 panels.}
\Description{Four scatter panels show Session~1 and Session~2 for the female-rates-male and male-rates-female directions. Each panel plots candidate models with utility, measured by the mean per-participant Pearson correlation with the liking score, on the vertical axis and redundancy, operationalized as the mean per-participant Pearson correlation with predictions from Claude, the prespecified transcript-only LLM, on the horizontal axis. In every panel, the large language models and multimodal large language models cluster on the right at prediction correlations above roughly 0.3. HuBERT, the supervised speech predictor, sits alone in the upper-left region, combining utility around 0.1 to 0.2 with prediction correlation near zero. Claude anchors the right edge at a correlation of 1.0. Arrows indicate the preferred direction for a fusion complement: upward for higher utility and leftward for lower prediction correlation.}
\label{fig:complementarity_landscape}
\end{figure*}

To assess the robustness of choosing Claude for transcript-only prediction and
HuBERT for supervised speech prediction
(\S4.3 in the main paper), we re-ran
weighted score-level late fusion with alternative supervised representatives:
Sentence-T5~\cite{ni-etal-2022-sentence}, BERT~\cite{devlin-etal-2019-bert},
and the Japanese Linguistic Inquiry and Word Count (J-LIWC)~\cite{Igarashi2022JLIWC} for text, and
openSMILE~\cite{Eyben2010OpenSMILE} for speech. Table~\ref{tab:substitution_ablation}
summarizes the resulting representative-selection and substitution analyses.
Figure~\ref{fig:complementarity_landscape} complements this ablation by
situating the full candidate pool in the utility--redundancy plane
referenced in \S4.3 in the main paper. Across all four conditions, HuBERT
occupies a favorable region among the speech candidates. It combines
relatively high utility with limited prediction correlation with Claude.

\begin{table}[t]
  \centering
    \caption{Per-participant $r$ for selection and substitution analyses of the transcript-only and speech predictors.}
    \label{tab:substitution_ablation}
    \Description{Rows compare Claude, the prespecified transcript-only LLM; its auxiliary thinking-enabled configuration; the main fusion with HuBERT, the supervised speech predictor; and substitutions for either predictor. Columns give per-participant r for the four session-by-direction conditions.}
    \small
    \setlength{\tabcolsep}{4pt}
    \renewcommand{\arraystretch}{1.05}
    \begin{tabular}{@{}lcccc@{}}
      \toprule
      Configuration & S1 F2M & S1 M2F & S2 F2M & S2 M2F \\
      \midrule
      Claude alone                        & .248 & .163 & .244 & .238 \\
      Claude (thinking) alone             & .246 & .165 & .250 & .202 \\
      Claude + HuBERT (main)              & $\mathbf{.281}$ & $\mathbf{.164}$ & $\mathbf{.323}$ & $\mathbf{.248}$ \\
      \addlinespace[2pt]
      \multicolumn{5}{@{}l}{\textit{Text substitution (HuBERT held fixed, cross-modal)}} \\
      BERT + HuBERT                       & .111 & .098 & .195 & .101 \\
      J-LIWC + HuBERT                     & .142 & .092 & .180 & .183 \\
      \addlinespace[2pt]
      \multicolumn{5}{@{}l}{\textit{Speech substitution (Claude held fixed, cross-modal)}} \\
      Claude + openSMILE                  & .248 & .163 & .239 & .223 \\
      \addlinespace[2pt]
      \multicolumn{5}{@{}l}{\textit{Within-modality comparison (text + text, same modality)}} \\
      Claude + Sentence-T5                & .228 & .163 & .306 & .238 \\
      Claude + BERT                       & .233 & .120 & .177 & .114 \\
      Claude + J-LIWC                     & .240 & .155 & .219 & .238 \\
      \bottomrule
    \end{tabular}
    \par\vspace{5pt}
    \begin{minipage}{\linewidth}
      \footnotesize
      \raggedright
      \textit{Note.} Fusion is the weighted score-level late fusion described in
        \S4.3 in the main paper. \textbf{Bold} marks the main Claude--HuBERT fusion.
    \end{minipage}
\end{table}

The thinking-enabled auxiliary Claude configuration has similar
per-participant $r$ to disabled Claude in three conditions and lower $r$ in
Session~2 M2F; it does not systematically improve per-participant $r$ over
disabled Claude. We therefore retain disabled Claude as the prespecified
transcript-only LLM in the main analysis.

Replacing Claude with any supervised text model lowers fusion $r$ relative to
the main Claude--HuBERT fusion in every condition. Claude's zero-shot
transcript-only prediction is therefore not interchangeable with a supervised
text prediction under this fusion setup. The three supervised text alternatives
rank in no stable order across conditions.

Replacing HuBERT with openSMILE lowers fusion $r$ in Session~2
($\Delta r = -.02$ to $-.09$, substituted fusion minus the main
Claude--HuBERT fusion). In Session~1, this substituted fusion matches Claude
alone ($r = .248, .163$), with the weight-selection procedure
discounting openSMILE almost entirely. This pattern is consistent with
openSMILE's standalone $r$ being undefined on the full Session~1
rater set; see the note to Table~1 in the main paper.

None of the tested within-modality (text + text) ensembles consistently
improves on Claude alone. The Claude--BERT
text--text combination is below Claude alone in all four conditions.
The Sentence-T5 combination approaches the main fusion only in Session~2 F2M
($.306$ vs.\ $.323$), while it underperforms in the other three conditions.
The main Claude--HuBERT pairing is the only tested combination whose $r$
numerically exceeds Claude alone in every condition; the tested text--text
ensembles do not reproduce this four-condition pattern.

\subsection{Per-Participant CCC for Fusion Components}
\label{sec:app_fusion_ccc}

Although the main paper focuses on per-participant $r$ as the primary metric,
CCC (concordance correlation coefficient)~\cite{Lin1989CCC} is used for supervised training and fusion-weight selection for the
calibration-sensitive reasons described in \S4.4 in the main paper. Table~\ref{tab:app_fusion_ccc} reports
the per-participant CCC for each fusion component (Claude, HuBERT, and their
weighted combination). Fusion CCC exceeds Claude's CCC in three of the four
conditions, showing qualitative alignment with the condition-level $r$ results.

\begin{table}[t]
  \centering
    \caption{Per-participant CCC for Claude, HuBERT, and their weighted score-level late fusion.}
    \label{tab:app_fusion_ccc}
    \Description{Rows list predictions from Claude, the prespecified transcript-only LLM; HuBERT, the supervised speech predictor; and their weighted score-level late fusion. Columns give per-participant concordance correlation coefficients for the four session-by-direction conditions.}
    \small
    \setlength{\tabcolsep}{4pt}
    \renewcommand{\arraystretch}{1.05}
    \begin{tabular}{@{}lcccc@{}}
      \toprule
      Component & S1 F2M & S1 M2F & S2 F2M & S2 M2F \\
      \midrule
      Claude       & $.081{\scriptscriptstyle \pm .016}$ & $\mathbf{.073}{\scriptscriptstyle \pm .016}$ & $.067{\scriptscriptstyle \pm .017}$ & $.078{\scriptscriptstyle \pm .017}$ \\
      HuBERT       & $.054{\scriptscriptstyle \pm .024}$ & $.057{\scriptscriptstyle \pm .024}$ & $.057{\scriptscriptstyle \pm .024}$ & $.084{\scriptscriptstyle \pm .025}$ \\
      Fusion       & $\mathbf{.101}{\scriptscriptstyle \pm .019}$ & $.071{\scriptscriptstyle \pm .017}$ & $\mathbf{.105}{\scriptscriptstyle \pm .019}$ & $\mathbf{.094}{\scriptscriptstyle \pm .023}$ \\
      \bottomrule
    \end{tabular}
    \par\vspace{5pt}
    \begin{minipage}{\linewidth}
      \footnotesize
      \raggedright
      \textit{Note.} \textbf{Bold} indicates the best CCC per column. $\pm$ values denote the standard error of the mean (SE) across participants.
    \end{minipage}
\end{table}

Fusion CCC exceeds Claude's CCC in Session~1 F2M, Session~2 F2M, and
Session~2 M2F. Here, $\Delta r$ and $\Delta\mathrm{CCC}$ denote Fusion minus
Claude for per-participant $r$ and CCC, respectively. Both changes are near
zero in Session~1 M2F, consistent with the near-zero fusion gain in $r$ for
that condition.

\subsection{Equal-Weight vs.\ Weighted Fusion}
\label{sec:app_equal_vs_weighted}

\S6.2 in the main paper examines weighted averaging as a way to reduce the
influence of a weak component rather than produce uniform improvement.
Table~\ref{tab:equal_vs_weighted} reports per-participant $r$ for the
main fusion under both equal-weight ($w = 0.5$) and fold-selected
weighted averaging.
For the primary weighted fusion, paired $\Delta r$ relative to Claude
was $+.034$, $+.001$, $+.079$, and $+.011$ for Session~1 F2M, Session~1 M2F,
Session~2 F2M, and Session~2 M2F, respectively; none reached significance after Holm correction
(adjusted $p=.11$, $1.00$, $.29$, and $1.00$; $m=4$).

\begin{table}[t]
  \centering
    \caption{Per-participant $r$ for equal-weight versus fold-selected weighted score-level late fusion of Claude and HuBERT.}
    \label{tab:equal_vs_weighted}
    \Description{Rows compare three configurations: Claude alone, as the prespecified transcript-only LLM; its equal-weight fusion with HuBERT, the supervised speech predictor; and their fold-selected weighted fusion. Columns give per-participant r for the four conditions.}
    \small
    \setlength{\tabcolsep}{4pt}
    \renewcommand{\arraystretch}{1.05}
    \begin{tabular}{@{}lcccc@{}}
      \toprule
      Method & S1 F2M & S1 M2F & S2 F2M & S2 M2F \\
      \midrule
      Claude (reference)              & .248 & .163 & .244 & .238 \\
      Equal-weight average ($w{=}0.5$) & .185 & .162 & .316 & .250 \\
      Weighted average (main)         & $\mathbf{.281}$ & $\mathbf{.164}$ & $\mathbf{.323}$ & $\mathbf{.248}$ \\
      \bottomrule
    \end{tabular}
    \par\vspace{5pt}
    \begin{minipage}{\linewidth}
      \footnotesize
      \raggedright
      \textit{Note.} Weights in the fold-selected Claude--HuBERT fusion are selected
        per fold via a 21-point grid search over the pooled held-out
        predictions of the remaining folds
        (\S4.3 in the main paper). \textbf{Bold} marks the fold-selected Claude--HuBERT fusion.
    \end{minipage}
\end{table}

In Session~2 F2M, where HuBERT has its highest individual $r$, equal-weight
averaging already captures most of the gain ($r{=}.316$ vs.\ weighted
$.323$). In Session~1 F2M, where HuBERT is weaker, equal-weight averaging
falls below Claude alone ($.185$ vs.\ $.248$), while the weighted
version down-weights HuBERT and recovers an $r$ of $.281$. In both M2F
conditions, equal-weight and weighted averaging are essentially tied.
These results suggest that the main role of weighted
averaging is robustness to a weak component rather than a uniform additive gain.

\begin{figure}[t]
\centering
\includegraphics[width=\columnwidth]{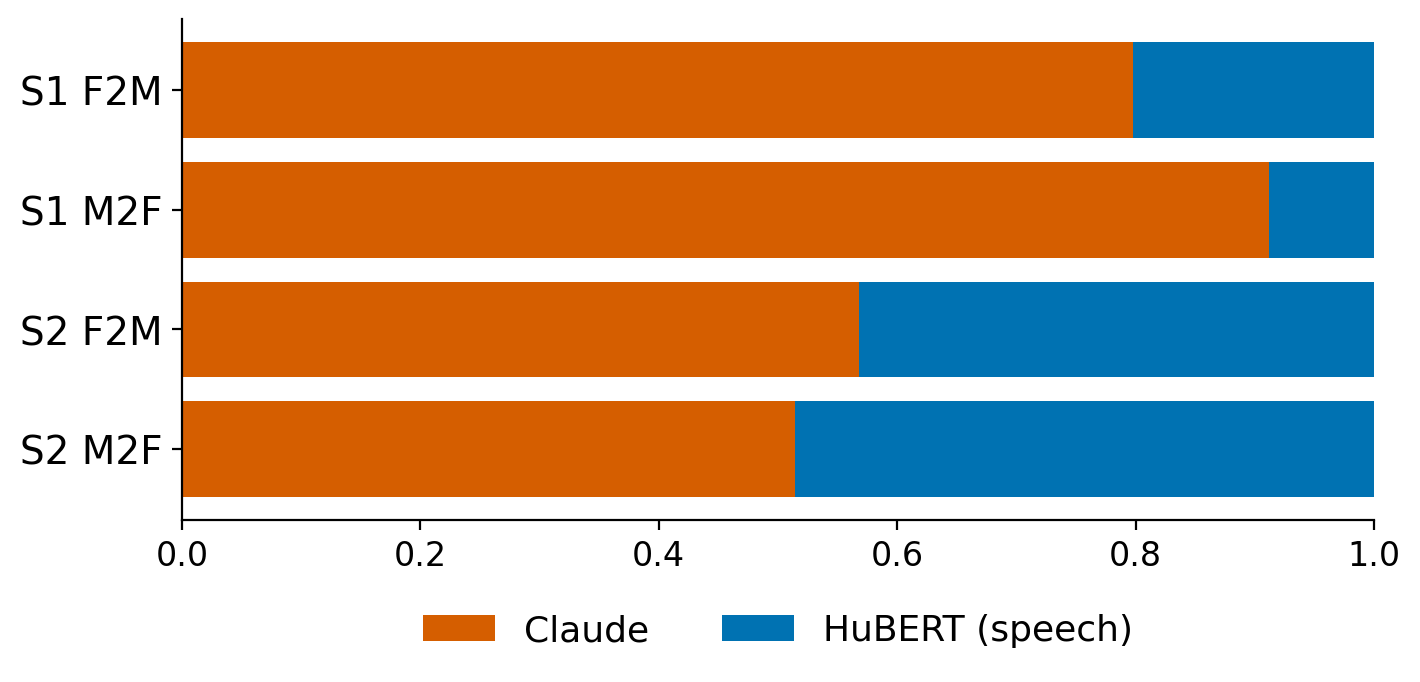}
\caption{Mean selected weights for the Claude transcript-only and HuBERT speech predictions across the 25 cross-validation folds, by condition.}
\label{fig:app_weights}
\Description{A horizontal stacked bar chart shows the shares of the Claude transcript weight and the HuBERT speech weight in the fused prediction. The Claude share is about 0.80 in Session~1 when female participants rate male partners, 0.91 in Session~1 when male participants rate female partners, 0.57 in Session~2 when female participants rate male partners, and 0.51 in Session~2 when male participants rate female partners. The remaining share in each condition goes to HuBERT.}
\end{figure}

Figure~\ref{fig:app_weights} shows the mean selected weights across
the 25 cross-validation folds. The selection procedure assigns HuBERT
comparable weight to the Claude branch in Session~2, where HuBERT has higher
individual $r$ than in Session~1. In Session~1, where HuBERT has lower
individual $r$ than in Session~2, the selection procedure concentrates on the Claude branch.
This pattern is consistent with the robustness-against-noise interpretation
above.

\subsection{Incremental Variance Decomposition}
\label{sec:app_incremental_variance}

\S6.1 in the main paper reports positive in-sample incremental $R^2$ for
each predictor given the other in all four conditions.
Table~\ref{tab:app_incremental_variance} provides the full decomposition.
For each participant with $\geq 3$ rated partners, we compute the within-rater
Pearson correlations between Claude prediction and ground truth ($r_C$), HuBERT
and ground truth ($r_H$), and the two predictions ($r_{CH}$). From these we
derive the multiple $R^2$ of the joint regression ($R^2_\text{joint}$) and
the incremental $R^2$ of each predictor given the other
($\Delta R^2_{H \mid C}$, $\Delta R^2_{C \mid H}$). We also report partial
correlations $r(\hat{y}_H, y \mid \hat{y}_C)$ and
$r(\hat{y}_C, y \mid \hat{y}_H)$, aggregated by Fisher-$z$ mean. The two
incremental $R^2$ contrasts are reported with 95\% participant-bootstrap
confidence intervals (CIs; 2{,}000 resamples).

\begin{table}[t]
  \centering
    \caption{Full within-rater variance decomposition of predictions from Claude and HuBERT.}
    \label{tab:app_incremental_variance}
    \Description{For each condition, the table reports the within-rater correlation between predictions from Claude, the prespecified transcript-only LLM, and HuBERT, the supervised speech predictor; their marginal and joint explained variance; their mutual incremental explained variance; and their shared explained variance.}
    \small
    \setlength{\tabcolsep}{3pt}
    \renewcommand{\arraystretch}{1.15}
    \begin{tabular}{@{}lcccc@{}}
      \toprule
      Metric & S1 F2M & S1 M2F & S2 F2M & S2 M2F \\
      \midrule
      $n$ raters                 & $73$  & $72$  & $75$  & $72$ \\
      \addlinespace[3pt]
      $r_C$                     & $+.30$ & $+.18$ & $+.29$ & $+.29$ \\
      $r_H$                     & $+.12$ & $+.13$ & $+.24$ & $+.21$ \\
      $r_{CH}$                  & $-.08$ & $+.13$ & $+.05$ & $+.07$ \\
      \addlinespace[3pt]
      $R^2_C$                   & $.23$ & $.17$ & $.24$ & $.20$ \\
      $R^2_H$                   & $.21$ & $.22$ & $.22$ & $.17$ \\
      $R^2_\text{joint}$        & $.40$ & $.37$ & $.45$ & $.34$ \\
      Shared $R^2$              & $.04$ & $.01$ & $.01$ & $.03$ \\
      \addlinespace[3pt]
      \multirow{2}{*}{$\Delta R^2_{H \mid C}$}
        & $+.17$ & $+.20$ & $+.21$ & $+.14$ \\[-2pt]
        & {\scriptsize$[.13,\,.22]$} & {\scriptsize$[.15,\,.26]$} & {\scriptsize$[.16,\,.26]$} & {\scriptsize$[.11,\,.17]$} \\
      \addlinespace[2pt]
      \multirow{2}{*}{$\Delta R^2_{C \mid H}$}
        & $+.19$ & $+.15$ & $+.24$ & $+.17$ \\[-2pt]
        & {\scriptsize$[.15,\,.24]$} & {\scriptsize$[.11,\,.19]$} & {\scriptsize$[.19,\,.29]$} & {\scriptsize$[.13,\,.21]$} \\
      \addlinespace[3pt]
      partial $r_{H \mid C}$     & $+.16$ & $+.12$ & $+.15$ & $+.21$ \\
      partial $r_{C \mid H}$     & $+.28$ & $+.21$ & $+.39$ & $+.28$ \\
      \bottomrule
    \end{tabular}
    \par\vspace{5pt}
    \begin{minipage}{\linewidth}
      \footnotesize
      \raggedright
      \textit{Note.} $C$ denotes the Claude transcript-only prediction, and $H$
        denotes the HuBERT speech prediction.
        Values are computed within rater and aggregated across participants
        (Fisher-$z$ means for correlations; arithmetic means for $R^2$).
        Shared $R^2 = R^2_C + R^2_H - R^2_\text{joint}$. The 95\% CIs,
        shown for the incremental $\Delta R^2$ rows, use 2{,}000
        participant-bootstrap resamples.
    \end{minipage}
\end{table}

The shared $R^2$ between Claude and HuBERT predictions is at most $.04$ across
the four conditions, small relative to each predictor's marginal
$R^2 \in [.17, .24]$. The two increments are comparable in magnitude, and neither
direction is uniformly larger: $\Delta R^2_{C \mid H}$ exceeds
$\Delta R^2_{H \mid C}$ in three of four conditions, with Session~1 M2F reversed.
This pattern is consistent with limited overlap between the two predictions.
However, these increments are in-sample and nonnegative by construction,
so their mere positivity does not demonstrate out-of-sample incremental
value. Within-rater $z$-scoring leaves these quantities unchanged, so
prediction scale compression does not affect the decomposition. Moreover,
weak overlap between the predictors cannot distinguish
genuinely distinct information from partially independent noise around the
same underlying signal. The decomposition therefore describes limited
overlap and comparable bidirectional increments in this sample, but does
not establish their mechanism or generalizability.

\subsection{Per-Participant vs.\ Global \texorpdfstring{$r$}{r}}
\label{sec:app_global_r}

Per-participant $r$ is adopted as the primary metric in \S4.4 in the main
paper. Table~\ref{tab:per_vs_global_r} complements the main results with the
corresponding \emph{global} $r$ (Pearson correlation over all pooled pairs)
to show where the two metrics diverge.

\begin{table}[t]
  \centering
    \caption{Per-participant $r$ versus global $r$ for Claude, HuBERT, and their fusion, by condition.}
    \label{tab:per_vs_global_r}
    \Description{Rows group the four conditions, with separate entries for mean per-participant r and pooled global r. Columns compare predictions from Claude, the prespecified transcript-only LLM; HuBERT, the supervised speech predictor; and their fusion.}
    \small
    \setlength{\tabcolsep}{4pt}
    \renewcommand{\arraystretch}{1.05}
    \begin{tabular}{@{}llccc@{}}
      \toprule
      Condition & Metric & Claude & HuBERT & Fusion \\
      \midrule
      \multirow{2}{*}{S1 F2M} & Per-participant $r$ & .248 & .105 & .281 \\
                              & Global $r$     & .241 & $-.043$ & .180 \\
      \addlinespace[2pt]
      \multirow{2}{*}{S1 M2F} & Per-participant $r$ & .163 & .092 & .164 \\
                              & Global $r$     & .114 & .112 & .126 \\
      \addlinespace[2pt]
      \multirow{2}{*}{S2 F2M} & Per-participant $r$ & .244 & .212 & .323 \\
                              & Global $r$     & .238 & .024 & .134 \\
      \addlinespace[2pt]
      \multirow{2}{*}{S2 M2F} & Per-participant $r$ & .238 & .183 & .248 \\
                              & Global $r$     & .099 & .154 & .170 \\
      \bottomrule
    \end{tabular}
\end{table}

In Session~2 F2M, per-participant $r$ for the fusion ($.323$) exceeds
Claude alone ($.244$), whereas global $r$ reverses this ordering (fusion
$.134$ vs.\ Claude $.238$). The reversal shows that pooled and within-participant
correlations can yield different model orderings. Per-participant $r$ measures
relative partner differentiation within each rater. Computing $r$ separately
for each rater prevents between-participant differences in score location and
scale from contributing to the metric. By contrast, global $r$ mixes within-
and between-participant covariation. The table alone does not identify which
between-participant component produces the reversal. Per-participant $r$ is
aligned with the matchmaking task (\S3.3 in the main paper) and is used as the
primary metric for that reason. The reversal illustrates the consequence of
the metric choice rather than establishing a fusion mechanism.

\subsection{Partner-Level Consensus in the Evaluation Target}
\label{sec:app_partner_consensus}

Per-participant $r$ does not distinguish rater-specific preference from
partners' generally shared appeal. To quantify the latter component in the
evaluation target, we construct a descriptive \emph{partner-level consensus
reference}. For each (rater, partner) pair within a session and conversation
group, the reference prediction is the mean liking score that the same partner
received from the other raters in that group (leave-one-out; three or four
donor ratings per pair in the analyzed set). We then compute Pearson $r$ within
each rater and average across raters, as in the primary evaluation. Because it
uses other raters' outcome labels, this reference is unavailable at inference
and serves as a diagnostic of the evaluation target, not as a deployable model
baseline. Table~\ref{tab:app_partner_consensus} compares the reference with the
main-fusion components to place their values on the same scale.

\begin{table}[t]
  \centering
    \caption{Partner-level consensus reference versus the Claude transcript-only prediction, HuBERT speech prediction, and their fusion. Per-participant $r$.}
    \label{tab:app_partner_consensus}
    \Description{Rows give the four conditions and their mean. Columns compare the partner-level consensus reference with predictions from Claude, the prespecified transcript-only LLM; HuBERT, the supervised speech predictor; their fusion; and the fusion-minus-reference difference.}
    \small
    \setlength{\tabcolsep}{3pt}
    \renewcommand{\arraystretch}{1.05}
    \begin{tabular}{@{}lccccc@{}}
      \toprule
      Condition & Consensus ref. & Claude & HuBERT & Fusion & \shortstack{$\Delta r$\\(Fusion $-$ Ref.)} \\
      \midrule
      S1 F2M & $.462$ & $.248$ & $.105$ & $.281$ & $-.180$ \\
      S1 M2F & $.362$ & $.163$ & $.092$ & $.164$ & $-.198$ \\
      S2 F2M & $.480$ & $.244$ & $.212$ & $.323$ & $-.157$ \\
      S2 M2F & $.384$ & $.238$ & $.183$ & $.248$ & $-.135$ \\
      \midrule
      Mean   & $.422$ & $.223$ & $.148$ & $.254$ & $-.168$ \\
      \bottomrule
    \end{tabular}
    \par\vspace{5pt}
    \begin{minipage}{\linewidth}
      \footnotesize
      \raggedright
      \textit{Note.} The reference uses other raters'
        ground-truth scores from the same session and conversation group;
        these labels are unavailable to a model at inference. Claude values use
        Claude Sonnet~4.6 with extended thinking disabled, and Fusion denotes its weighted score-level late fusion
        with HuBERT. $\Delta r$ (Fusion $-$ Ref.) is the Fusion value minus
        the consensus-reference value, computed before rounding. Model values are
        reproduced from Table~1 in the main paper. Mean is the arithmetic
        mean of the four conditions.
    \end{minipage}
\end{table}

The reference exceeds the fusion in every condition, by $.135$--$.198$ in
per-participant $r$. This gap is not evidence that the reference is a better
predictive model, because the comparison is not information-matched: the
reference directly accesses other raters' outcome labels, whereas the models
receive conversation data. Instead, the result indicates that within-rater
partner ordering contains a substantial group-level consensus component in
received liking; the present evaluation does not isolate uniquely
rater-specific matching.

\subsection{Fusion Robustness to LLM Choice: Claude vs.\ GPT-5.4}
\label{sec:app_gpt_robustness}

The main fusion (\S4.3 in the main paper) uses Claude as the prespecified
transcript-only LLM. GPT-5.4 is the other transcript-only LLM showing
competitive per-participant $r$ in Session~1 (Table~1 in the main paper). To verify that the
condition-dependent $\Delta r$ pattern is not driven by the specific LLM,
Table~\ref{tab:fusion_robustness_llm} compares the corresponding HuBERT
fusions. The evaluation protocol, hyperparameter settings, and
evaluated participant set are held fixed; fusion weights are selected
separately for each underlying LLM using the same procedure.

\begin{table}[t]
  \centering
    \caption{Fusion robustness to transcript-only LLM choice, with each LLM paired with HuBERT.}
    \label{tab:fusion_robustness_llm}
    \Description{Rows give the four conditions. Columns compare fusion of the HuBERT speech prediction with either Claude or GPT-5.4 as the transcript-only LLM, and the corresponding fusion-minus-LLM gains.}
    \small
    \setlength{\tabcolsep}{4pt}
    \renewcommand{\arraystretch}{1.05}
    \begin{tabular}{@{}lcccc@{}}
      \toprule
       & \multicolumn{2}{c}{\textit{Fusion $r$}} & \multicolumn{2}{c}{\textit{$\Delta r$ (Fusion $-$ LLM)}} \\
      \cmidrule(lr){2-3}\cmidrule(lr){4-5}
      Condition & Claude & GPT-5.4 & Claude & GPT-5.4 \\
      \midrule
      S1 F2M & $.281{\scriptscriptstyle \pm .049}$ & $.295{\scriptscriptstyle \pm .047}$ & $+.034{\scriptscriptstyle \pm .027}$ & $+.039{\scriptscriptstyle \pm .027}$ \\
      S1 M2F & $.164{\scriptscriptstyle \pm .046}$ & $.203{\scriptscriptstyle \pm .052}$ & $+.001{\scriptscriptstyle \pm .010}$ & $.000{\scriptscriptstyle \pm .016}$ \\
      S2 F2M & $\mathbf{.323}{\scriptscriptstyle \pm .044}$ & $\mathbf{.309}{\scriptscriptstyle \pm .049}$ & $\mathbf{+.079}{\scriptscriptstyle \pm .053}$ & $\mathbf{+.114}{\scriptscriptstyle \pm .050}$ \\
      S2 M2F & $.248{\scriptscriptstyle \pm .044}$ & $.255{\scriptscriptstyle \pm .044}$ & $+.011{\scriptscriptstyle \pm .042}$ & $+.017{\scriptscriptstyle \pm .043}$ \\
      \bottomrule
    \end{tabular}
    \par\vspace{5pt}
    \begin{minipage}{\linewidth}
      \footnotesize
      \raggedright
      \textit{Note.} $\Delta r$ is the paired per-participant
        difference between Fusion and the underlying LLM. $\pm$: standard
        error of the mean across participants. \textbf{Bold} marks the
        largest value per column.
    \end{minipage}
\end{table}

The condition ranking of fusion $r$ matches between the two transcript-only LLMs
(Session~2 F2M highest, Session~1 M2F lowest), and the per-condition differences are
small relative to the reported standard errors. The $\Delta r$ pattern also
replicates descriptively: both LLMs
yield the largest gain in Session~2 F2M and a null gain in Session~1 M2F. The Session~2 F2M
$\Delta r$ is larger for GPT-5.4 ($+.114$) than for Claude
($+.079$), consistent with HuBERT having more room to
contribute when the paired LLM's own $r$ is lower ($.196$ vs.\ $.244$). Thus, the
condition-level pattern is reproduced with one alternative LLM and is not
unique to the Claude configuration, although this comparison does not
establish model-independent generality.

\subsection{Late vs.\ Early Fusion: Architectural Comparison}
\label{sec:app_late_vs_early}

Because the main Claude--HuBERT pair can only be fused at score level, the
supervised Sentence-T5 + HuBERT pair provides the only setting in which we can
compare score-level late fusion with an utterance-level early-fusion
alternative. Table~1 in the main paper reports both variants for this
supervised text--speech pair. Late fusion yields nearly the same $r$ as early
fusion in Session~1 F2M ($.103$ vs.\ $.106$) and has higher $r$ in the other
three conditions. The weighted score-level late fusion assigns
$w_\text{T5} \approx 0$ in nearly all folds, effectively discarding
the weak Sentence-T5 branch through an explicit branch-level scalar weight.
The early-fusion projection can in principle attenuate text dimensions,
but it has no directly equivalent scalar gate because the 768-dimensional
text channel is concatenated with HuBERT at every utterance. The observed
comparison does not by itself identify why early fusion has lower $r$ in
three conditions.

\paragraph{Early-fusion architecture.}
Sentence-T5 (768-d) and HuBERT (1024-d) are concatenated at each utterance
into a 1{,}792-dimensional stream. The stream is fed to a shared attention
pooling + regression head, trained under the same settings as the
single-branch baselines.

\subsection{External Human Annotation Check}
\label{sec:app_human_annotation}

To illustrate the distinction between rank consistency and absolute score
agreement described in \S4.4 in the main paper, three external annotators
independently viewed 50 overhead video
recordings with synchronized audio (25 conversations from each session) and
estimated the post-conversation liking scores in both directions. For each
conversation--direction case, each annotator completed the same 13 items used
for the prediction target on the 1--9 scale. The annotated subset comprised
cross-validation fold 17 in Session~1 and fold 15 in Session~2, yielding 100
conversation--direction cases; it was not used for model training or model
selection.

Agreement on the sum of the 13 items, as measured by the intraclass
correlation coefficient (ICC), was low for absolute agreement,
ICC(2,$k$) $=.29$ (95\% CI $[-.06,.57]$), but higher for consistency,
ICC(3,$k$) $=.69$ (95\% CI $[.57,.78]$). Pairwise Pearson correlations
between annotators ranged from $.53$ to $.61$. This pattern is consistent
with substantial differences in scale placement alongside moderate agreement
in relative ordering, illustrating why the main evaluation reports both a
relative-differentiation primary metric and a calibration-sensitive secondary metric.
Given the small diagnostic subset, we do not treat these annotations as a
human-performance benchmark.

\section{Evaluation and Implementation Details}
\label{sec:app_reproducibility}

\subsection{Corpus Subset and Recording Details}
\label{sec:app_corpus_descriptives}

Because sample composition and session setup affect the interpretation of the
condition-level results, this section documents dataset details that are
summarized more briefly in \S3.1 and \S3.2 of the main paper. It reports
condition-level liking-score descriptives, explains why each session contains
624 rather than 625 conversations, and clarifies how the nominal session
durations were handled. Per-condition means and standard deviations, computed
after applying the rater exclusion rule, are:
Session~1 F2M $M = 4.36$ ($SD = 1.50$), M2F $M = 4.77$ ($SD = 1.47$);
Session~2 F2M $M = 4.49$ ($SD = 1.53$), M2F $M = 4.94$ ($SD = 1.47$).

\paragraph{Conversation count.}
Each session contains 624 rather than the nominal $25 \times 25 = 625$
pairwise conversations: one male--female pair has two recorded
conversations in each session, and only the earlier recording (by
timestamp) is retained for that pair.

\paragraph{Session duration.}
We use the session-specific recordings as provided, with nominal durations of
5 minutes for Session~1 and 10 minutes for Session~2. Neither transcript-window construction
for the LLMs nor supervised preprocessing applies an additional hard timestamp
cutoff.

\subsection{Cross-Validation Protocol Details}
\label{sec:app_cv_protocol}

Because participants can appear in more than one conversation group, the
handling of participant identities and outcome labels across training,
validation, and model selection is central to assessing leakage control and
the scope of participant separation. \S4.4 in the main paper summarizes the
25-fold Leave-One-Group-Out (LOGO) protocol; this section specifies
validation-fold selection and training-side exclusions, while
Section~\ref{sec:app_weight_sensitivity} separately examines participant
overlap in fusion-weight selection.

\paragraph{Validation fold.}
For each test fold, one fold that shares no participants with the test fold
is randomly selected as the validation fold using the fixed random seed 42.
The validation fold is used for early stopping (patience~20 epochs) and does
not contribute to the final test metrics.

\paragraph{Training-side exclusions.}
From the training set we further exclude (a) participants with fewer than
three partners in the given rating direction, and (b) low-variance
participants whose liking-score SD falls below the median participant SD
minus 2.5 times its median absolute deviation (MAD). These exclusions avoid
unstable participant-level CCC estimates from participants with few partners
or near-constant ratings. They apply only to training; the test and validation
evaluation sets are unchanged.

\paragraph{Late fusion weight optimization.}
As described in \S4.3 in the main paper, the fusion weight $w$ is selected
per test fold on the pooled held-out predictions of the remaining folds,
and the selected $w$ is applied to the held-out test fold without further
tuning.

\subsection{Sensitivity to Participant Overlap in Fusion-Weight Selection}
\label{sec:app_weight_sensitivity}

The fusion weight $w$ is selected per test fold on the pooled held-out
predictions of the remaining folds (\S4.3 in the main paper). Because
participants can appear in more than one conversation group, pairs
involving test-fold participants make up approximately 9\% of this
selection pool, so the pool is not strictly participant-disjoint from
the test fold. To quantify the effect of this coupling, we repeated the
weight selection after removing every pair that involves a test-fold
participant in either role (leaving 91\% of the pool) and re-derived
all fusion results.

Table~\ref{tab:app_weight_sensitivity} compares the two protocols. All
conclusions reported in the main paper hold under both. Pairwise accuracy (PW)
improves over Claude alone in all four conditions (Holm-corrected
$p < .05$). The fused model exceeds HuBERT alone in PW in
the same three conditions, with Session~1 M2F remaining the exception.
The Spearman association between HuBERT's per-participant $r$ and fusion gain
$\Delta r$ stays in the $.70$--$.80$ range, with its largest value in
Session~2 F2M, and no paired $\Delta r$ reaches Holm-corrected
significance. The visible effect is confined to fusion $r$ in Session~1
(from $.281$ to $.262$ in F2M and from $.164$ to $.139$ in M2F). In these
Session~1 conditions, HuBERT is weak, the selected weight concentrates on
the Claude branch, and the observed change is consistent with greater sensitivity to
the composition of the selection pool. Session~2 values are essentially
unchanged.

\begin{table}[t]
  \centering
    \caption{Claude--HuBERT fusion results under pooled versus strictly participant-disjoint weight selection. $\Delta$PW denotes Fusion $-$ Claude, and Coupling $\rho$ denotes the Spearman correlation between HuBERT's per-participant $r$ and fusion gain $\Delta r$.}
    \label{tab:app_weight_sensitivity}
    \Description{Rows give the four conditions. Paired columns compare pooled and strictly participant-disjoint weight-selection protocols for fusion r, pairwise-accuracy gain over Claude, and the Spearman association between HuBERT's per-participant r and fusion gain in r relative to Claude. Claude is the prespecified transcript-only LLM, and HuBERT is the supervised speech predictor.}
    \small
    \setlength{\tabcolsep}{4pt}
    \renewcommand{\arraystretch}{1.05}
    \begin{tabular}{@{}lcccccc@{}}
      \toprule
       & \multicolumn{2}{c}{\textit{Fusion $r$}} & \multicolumn{2}{c}{\textit{$\Delta$PW (vs.\ Claude)}} & \multicolumn{2}{c}{\textit{Coupling $\rho$}} \\
      \cmidrule(lr){2-3}\cmidrule(lr){4-5}\cmidrule(lr){6-7}
      Condition & Pooled & Disjoint & Pooled & Disjoint & Pooled & Disjoint \\
      \midrule
      S1 F2M & $.281$ & $.262$ & $+.054^{***}$ & $+.046^{**}$ & $.76$ & $.73$ \\
      S1 M2F & $.164$ & $.139$ & $+.025^{**}$  & $+.017^{*}$  & $.70$ & $.70$ \\
      S2 F2M & $.323$ & $.321$ & $+.049^{*}$   & $+.050^{*}$  & $.80$ & $.80$ \\
      S2 M2F & $.248$ & $.250$ & $+.049^{*}$   & $+.052^{*}$  & $.72$ & $.70$ \\
      \bottomrule
    \end{tabular}
    \par\vspace{5pt}
    \begin{minipage}{\linewidth}
      \footnotesize
      \raggedright
      \textit{Note.} Pooled = weight selection on the pooled held-out
        predictions of the remaining folds (the protocol used throughout
        the paper); Disjoint = pairs involving test-fold participants
        additionally removed from the selection pool. $\Delta$PW is the
        aggregate pairwise-accuracy gain of the fusion over Claude alone;
        Coupling $\rho$ is the Spearman correlation between HuBERT's
        per-participant $r$ and the per-participant fusion gain $\Delta r$.
        $^{*}p<.05$, $^{**}p<.01$, $^{***}p<.001$ (Wilcoxon, Holm-corrected,
        $m=4$).
    \end{minipage}
\end{table}

\subsection{LLM Prompt Details}
\label{sec:app_prompt}

Because transcript-only LLM predictions depend on input scope, rater role, and
output constraints, this section documents what information each model received
and how the 13 item scores were elicited. \S4.2 in the main paper summarizes
the prompt design. The system and user
prompts are documented below, followed by an English translation and
the placeholder substitutions. The released code contains the exact
executable Japanese templates, including the full Q1--Q13 wording; the
item list is marked as omitted in the abridged display below. These items are
the Japanese translation of Rubin's Liking Scale~\cite{Ishii2023INTERACT}; see
\S3.2 in the main paper. The primary Claude configuration uses temperature 0
and forced tool use returning thirteen integer scores via a JSON schema.

\paragraph{System prompt (abridged original Japanese).}
\begin{CJK*}{UTF8}{min}
\begin{quote}\small
あなたはスピードデーティングの実験参加者をシミュレートしています。

状況: あなたは\{evaluator\_role\}参加者として、\{target\_role\}の相手と\{duration\}分間の対話をしたところです。以下にその対話の書き起こしが提示されます。

タスク: 対話の書き起こしの内容のみに基づいて、相手の\{target\_role\}に対するあなたの印象を、以下の13の質問それぞれについて1〜9の整数で回答してください。

回答尺度: 1 = 全くそう思わない; 2 = そう思わない; 3 = あまりそう思わない; 4 = どちらかといえばそう思わない; 5 = どちらでもない; 6 = どちらかといえばそう思う; 7 = ややそう思う; 8 = そう思う; 9 = 非常にそう思う。

\textit{[Q1--Q13 omitted from this abridged display.]}

すべての質問に必ず回答してください。対話の内容から判断できない場合は、対話の雰囲気や文脈から最も妥当と思われるスコアをつけてください。
\end{quote}
\end{CJK*}

\paragraph{User prompt (original Japanese).}
\begin{CJK*}{UTF8}{min}
\begin{quote}\small
以下は、あなた（\{evaluator\_role\}）と相手（\{target\_role\}）の対話の書き起こしです。\\[2pt]
\{transcript\}\\[2pt]
上記の対話に基づいて、13の質問項目すべてに回答してください。
\end{quote}
\end{CJK*}

\paragraph{System prompt (abridged English translation).}
\begin{quote}\small
You are simulating a speed-dating study participant.

Context: You have just completed a \texttt{\{duration\}}-minute conversation
with a \texttt{\{target\_role\}} partner as the \texttt{\{evaluator\_role\}}
side. The transcript of that conversation is provided below.

Task: Based solely on the transcript, rate your impression of the
\texttt{\{target\_role\}} partner on each of the 13 items below using an
integer between 1 and 9.

Rating scale: 1 = strongly disagree; 2 = disagree; 3 = somewhat disagree;
4 = mildly disagree; 5 = neither agree nor disagree; 6 = mildly agree;
7 = somewhat agree; 8 = agree; 9 = strongly agree.

\textit{[Q1--Q13 omitted from this abridged translation.]}

Please answer every question. If the transcript does not give a clear
signal, use the atmosphere and context of the conversation to assign the
most plausible score.
\end{quote}

\paragraph{User prompt (English translation).}
\begin{quote}\small
Below is the transcript of your conversation (\texttt{\{evaluator\_role\}})
with your partner (\texttt{\{target\_role\}}).\\[2pt]
\texttt{\{transcript\}}\\[2pt]
Based on this transcript, please answer all thirteen items.
\end{quote}

\paragraph{Transcript format.}
\begin{CJK*}{UTF8}{min}
Lines of the form \texttt{[MM:SS] 男性: utterance} or
\texttt{[MM:SS] 女性: utterance} (``男性'' = male, ``女性'' = female) are
concatenated in chronological order for the entire session. The complete
chronological transcript is provided to each LLM, including utterances from
both the rater and the partner; unlike the supervised branches, no rater-side
filtering is applied.
\end{CJK*}

\paragraph{Placeholder substitution.}
\begin{CJK*}{UTF8}{min}
\begin{center}
  \small\setlength{\tabcolsep}{6pt}
  \begin{tabular}{l|ll}
    \toprule
    Variable & F2M & M2F \\
    \midrule
    \texttt{\{evaluator\_role\}} & 女性 & 男性 \\
    \texttt{\{target\_role\}}    & 男性 & 女性 \\
    \bottomrule
  \end{tabular}
  \quad
  \begin{tabular}{l|cc}
    \toprule
    Variable & Session~1 & Session~2 \\
    \midrule
    \texttt{\{duration\}} & 5 & 10 \\
    \bottomrule
  \end{tabular}
\end{center}
\end{CJK*}

For transcript-only input, the compared LLMs use the same base prompt
(Claude Sonnet~4.6, GPT-5.4, Gemini~2.5 Flash, and Gemma~3 12B-IT), with
the output format adapted to each API's JSON schema, tool, or structured-output
mechanism. For text-plus-audio input, we use Gemini~2.5 Flash and
GPT-audio-mini; the input-description sentence and task instruction additionally
refer to the supplied audio. The exact variants are included in the released
code.

\subsection{openSMILE Feature Specification}
\label{sec:app_opensmile_features}

openSMILE appears as a substitution baseline for the speech branch, so the
feature inventory is included to define what information that baseline
represents. The openSMILE speech baseline in Table~1 in the main paper and in
Table~\ref{tab:substitution_ablation} uses a 338-dimensional feature vector
extracted per utterance with the openSMILE toolkit. We use the emobase
configuration for the functionals, restricted to eight low-level descriptors
(LLDs): F0, F0 envelope, line spectral pair frequencies, MFCC, PCM
intensity, PCM loudness, PCM zero-crossing rate, and voicing probability.
Each LLD is summarized by standard statistical functionals (mean, standard
deviation, range, extrema, linear-regression coefficients and error,
skewness, kurtosis), yielding 338 features in total.

\section{Exploratory Analyses and Extensions}
\label{sec:app_exploratory_extensions}

\subsection{Candidate Indicators of Fusion Gain}
\label{sec:app_null}

The coupling analysis in the main paper shows where fusion helps, but not
whether simple participant or linguistic descriptors identify those cases.
We include this null search to test such post hoc indicators and to avoid
over-interpreting fusion gain as evidence of a specific linguistic mechanism.
For each rater, the mean and SD of the true liking scores were correlated with fusion gain
$\Delta r$ (Fusion $-$ Claude) via Spearman correlation. In Session~2, SD
yielded $\rho=-.20$ (F2M) and $+.04$ (M2F), while the mean yielded
$\rho=+.09$ (F2M) and $-.05$ (M2F); all Holm-corrected $p>.16$, with
correction applied to the two descriptors within each direction ($m=2$).

In Session~2, we also tested 69 J-LIWC categories for both the
rater and partner (138 tests per direction), applying Benjamini--Hochberg
false discovery rate (FDR) correction at $q=.05$ within each direction. No association survived
correction. A nominal association between partner personal-pronoun frequency
and fusion gain in Session~2 F2M ($\rho=+.29$, uncorrected $p=.011$) did
not replicate in M2F ($\rho=-.09$). We therefore found no direction-robust
candidate indicator and do not interpret these exploratory associations as
evidence of a linguistic mechanism or statistical independence.
This null search is discussed in \S6.1 in the main paper.

\subsection{Visual Modality Extension (V-JEPA 2.1)}
\label{sec:app_vjepa}

As noted in \S6.3 in the main paper, the main analysis is restricted to text
and speech. To check whether available video qualitatively alters the reported
pattern, we report a preliminary visual extension using each rater's bust-up
video encoded locally with frozen V-JEPA~2.1
ViT-Giant~\cite{MurLabadia2026VJEPA21}; video is not sent to an external API.
Frames are sampled at 1~fps and grouped into 64-frame clips. Average pooling
over spatial tokens produces a 1{,}408-dimensional temporal embedding, which
is passed through the same rater-side attention-pooling and regression
architecture as the supervised branches in \S4.1 of the main paper.
Table~\ref{tab:app_vjepa_extension} reports the resulting unimodal, bimodal,
and trimodal performance.

\begin{table}[h]
  \centering
    \caption{Preliminary extension of the Claude--HuBERT fusion with V-JEPA~2.1. Per-participant $r$.}
    \label{tab:app_vjepa_extension}
    \Description{Rows compare a V-JEPA 2.1 visual prediction, the main fusion of Claude transcript-only and HuBERT speech predictions, and its extension with the V-JEPA 2.1 visual prediction. Columns give per-participant r for the four conditions.}
    \small
    \setlength{\tabcolsep}{1.5pt}
    \renewcommand{\arraystretch}{1.05}
    \begin{tabular}{@{}lcccc@{}}
      \toprule
      Configuration & S1 F2M & S1 M2F & S2 F2M & S2 M2F \\
      \midrule
      V-JEPA~2.1 alone                  & $-.022$ & $.120$ & $.203$ & $-.004$ \\
      Claude + HuBERT (main bimodal)             & $.281$ & $.164$ & $.323$ & $.248$ \\
      Claude + HuBERT + V-JEPA~2.1               & $.233$ & $.165$ & $.359$ & $.250$ \\
      \bottomrule
    \end{tabular}
\end{table}

Relative to the main bimodal result, the trimodal result decreases in
Session~1 F2M, increases by only $.001$--$.002$ in the two M2F conditions,
and increases by $.036$ in Session~2 F2M. Because this branch is preliminary and lies outside
the speech-focused scope of the main analysis, we do not interpret its
condition-specific pattern further.

\balance
\bibliographystyle{ACM-Reference-Format}
\bibliography{appendix-references}